\documentclass[11pt]{article}
\usepackage[square,sort,comma,numbers]{natbib}
\usepackage{amsmath, amssymb, amscd, amsthm, amsfonts}
\usepackage{graphicx}
\usepackage{hyperref}
\usepackage[utf8]{inputenc} 
\usepackage[T1]{fontenc}    
\usepackage{hyperref}       
\usepackage{url}            
\usepackage{booktabs}       
\usepackage{amsfonts}       
\usepackage{nicefrac}       
\usepackage{microtype}      
\usepackage[dvipsnames]{xcolor}
\usepackage{subfigure}      

\newtheorem{corollary}{Corollary}
\newtheorem{lemma}{Lemma}
\newtheorem{definition}{Definition}

\DeclareMathOperator*{\argmin}{arg\,min}

\newtheorem{theorem}{Theorem}
\usepackage{graphicx,float} 
\usepackage{amssymb}
\usepackage{algorithm}
\usepackage{algorithmic}
\usepackage{amsmath,bm}
\usepackage{makecell}

\usepackage{adjustbox}
\usepackage{balance}
\usepackage{mathtools}
\usepackage{multirow}

\usepackage{footnote}
\usepackage{etoolbox}
\makeatletter
\patchcmd{\@makecaption}
  {\scshape}
  {}
  {}
  {}
\makeatother
\usepackage{xcolor}

\usepackage{graphicx,float} 
\usepackage{makecell}

\usepackage{booktabs}
\usepackage{lipsum}
\usepackage{multicol}
\usepackage[capitalize,noabbrev]{cleveref}

\theoremstyle{plain}
\theoremstyle{definition}
\theoremstyle{remark}

\makeatletter

\title{Barycentric-alignment and Reconstruction Loss Minimization for Domain Generalization}

\author{Boyang Lyu$^{1 }$, Thuan Nguyen$^{1,2 }$, Prakash Ishwar$^3$, Matthias Scheutz$^2$, Shuchin Aeron$^1$
  \thanks{Author affiliations  \newline $^1$ - Tufts University, Dept. of ECE, $^2$ - Tufts University, Dept. of CS., $^3$ - Boston University, Dept. of ECE  \newline 
 \textbf{Corresponding authors}: Boyang Lyu, email: \url{Boyang.Lyu@tufts.edu}.} 
  } 
\date{}
\begin{document}
\maketitle

\begin{abstract}
This paper advances the theory and practice of Domain Generalization (DG) in machine learning. We consider the typical DG setting where the hypothesis is composed of a representation mapping followed by a labeling function. Within this setting, the majority of popular DG methods aim to jointly learn the representation and the labeling functions by minimizing a well-known upper bound for the classification risk in the unseen domain. In practice, however, methods based on this theoretical upper bound ignore a term that cannot be directly optimized due to its dual dependence on both the representation mapping and the unknown optimal labeling function in the unseen domain. To bridge this gap between theory and practice, we introduce a new upper bound that is free of terms having such dual dependence, resulting in a fully optimizable risk upper bound for the unseen domain. Our derivation leverages classical and recent \textit{transport inequalities} that link optimal transport metrics with information-theoretic measures. Compared to previous bounds, our bound introduces two new terms: {(i) the Wasserstein-2 barycenter term that aligns distributions between domains, and (ii) the reconstruction loss term that assesses the quality of representation in reconstructing the original data.} Based on this new upper bound, we propose a novel DG algorithm {named Wasserstein Barycenter Auto-Encoder (WBAE)} that simultaneously minimizes the classification loss, the barycenter loss, and the reconstruction loss. {Numerical results demonstrate that the proposed method outperforms current state-of-the-art DG algorithms on several datasets.}
\end{abstract}

\section{INTRODUCTION}\label{sec:intro}

Modern machine learning applications often encounter the problem that the training (seen) data and the test (unseen) data have different distributions, which can cause a deterioration in model performance.
For example, a model trained on data from one hospital may not work well when the test data is from another hospital \cite{gulrajani2020search}, a drowsiness driving estimator trained on one group of subjects may not perform well for other subjects \cite{cui2019eeg}, or a cognitive workload estimator based on fNIRS (functional near-infrared spectroscopy) measurements may not generalize well across sessions and subjects \cite{Lyu2021}. 
Methods that aim to mitigate this problem are broadly classified into two categories, namely Domain Adaptation (DA) \cite{ben2007analysis} and Domain Generalization (DG) \cite{blanchard2011generalizing}.
Both DA and DG aim to find a model that can generalize well in scenarios when the training data from the seen domain does not share the same distribution as the test data from the unseen domain. The key difference between DA and DG is that DA allows access to the (unlabeled) unseen domain data during the training process whereas DG does not. This makes DG a more challenging but more practical problem.

To address the problem of DG, motivated by the seminal theoretical works of \cite{ben2007analysis,ben2010theory}, the hypothesis is typically expressed as the composition of a representation function followed by a labeling function, \textit{e.g.}, see  \cite{albuquerque2019generalizing,dou2019domain,li2018domain,zhou2020domain}, and the representation and labeling functions are learned by minimizing an upper bound for the classification risk in the unseen domain derived in \cite{ben2007analysis,ben2010theory}. The upper bound 
consists of three terms: (1) the prediction risk on the mixture of seen domains, (2) the discrepancy or divergence between the data distributions of different domains in the representation space, and (3) a \textit{combined risk} across all domains that implicitly depends on both the representation mapping and the unknown optimal labeling function from the unseen domain. However, most current approaches disregard this dual dependency and treat the third term (\textit{combined risk}) as a constant while developing their algorithms. In fact, the majority of prominent works in DG and DA such as \cite{li2018domain,ganin2016domain,zhao2019learning} are essentially variations of the following strategy: ignore the \textit{combined risk} term and learn a domain-invariant representation mapping \textit{or} align the domains in the representation space, together with learning a common labeling function controlling the prediction loss across the seen domains. However, the \textit{combined risk} term is, in fact, a function of the representation mapping and should somehow be accounted for within the optimization process. Additional details of the shortcomings of previous upper bounds are provided in Appendix~\ref{apd: discussion subtlety}. 

To address these limitations, we revisit the analysis in \cite{ben2007analysis,ben2010theory} and derive a new upper bound that is free of terms with the dual dependence mentioned above. Our new bound consists of four terms: (1) the prediction risk across seen domains in the input space; (2) the discrepancy/divergence between the induced distributions of seen and unseen domains in the representation space, which can be approximated via the Wasserstein-2 barycenter \cite{OTAM} of seen domains; (3) the reconstruction loss term that measures how well the input can be reconstructed from its representation; and (4) a combined risk term that is independent of the representation mapping and labeling function to be learned. Our new bound differs from previous ones in two aspects. Firstly, it introduces two new terms: (a) the Wasserstein-2 barycenter term for domain alignment and (b) the reconstruction loss term for assessing the quality of representation in reconstructing the original data. 
We note that the Wasserstein-2 barycenter term for controlling the domain discrepancy in our bound is built in the representation space, which is better aligned with the practical implementation than previous Wasserstein-based bounds that are built in the data space.
Secondly, the combined risk in our bound is independent of the representation mapping and thus can be ignored during the optimization. Motivated by these theoretical results, we propose an Auto-Encoder-based model that interacts with the Wasserstein barycenter loss to achieve domain alignment.  

The contributions of this work can summarized as follows:
\begin{enumerate}
    \item {\textit{Contributions to Theory}: We propose a new upper bound for the risk of the unseen domain using classical and recent transport inequalities that link optimal transport metrics with information-theoretic measures. All terms in our new upper bound are optimizable in practice which overcomes the limitations of previous works and bridges the gap between previous theory and practice.}

    \item {\textit{Contributions to Algorithm Development and Practice}: We develop a novel algorithm for domain generalization based on our new upper bound. Our algorithm optimizes a new term that controls the domain discrepancy through Wasserstein-2 barycenter. Unlike previous Wasserstein distance-based bounds that form the domain discrepancy term in the data space but optimize it in the representation space, our domain discrepancy term is constructed and optimized in the representation space, making our practical implementation better aligned with the theory.} 

    \item {\textit{Gains over state-of-the-art methods}: Our algorithm consistently outperforms other theory-guided methods on PACS, VLCS, Office-Home, and TerraIncognita datasets, with a noticeable improvement of $1.7-2.8$ percentage points on average across all datasets.}
\end{enumerate}

\section{Related Work}
Our work falls within the DG framework wherein domain-invariant features are learned by decomposing the prediction function into a representation mapping followed by a labeling function. A recent example of this framework is \cite{albuquerque2019generalizing}, where the authors propose a three-part model consisting of a feature extractor, a classifier, and domain discriminators. The feature extractor learns the task-sensitive, but domain-invariant features via minimizing the cross-entropy loss with respect to the task label and maximizing the sum of domain discriminator losses. The domain discriminator loss is based on an estimate of the $\mathcal{H}$-divergence between all seen domains \cite{ben2010theory} and has roots in the works \cite{ganin2016domain,li2018deep} on Domain Adaptation. Following a similar idea, the authors of \cite{li2018domain} align the representation distributions from different domains by minimizing their Maximum Mean Discrepancy. In \cite{dou2019domain}, the authors adopt a gradient-based episodic training scheme for DG in which the extracted features are driven to simultaneously preserve global class information and local task-related clusters across seen domains by minimizing an alignment loss comprising soft class confusion matrices and a contrastive loss. The authors of \cite{arjovsky2019invariant} propose the Invariant Risk Minimization algorithm to learn features such that the optimal classifiers are matched across domains.
In \cite{nam2021reducing}, 
DG is achieved by disentangling style variation across domains from learned features. Among the large body of works on the DG problem, we regard \cite{ganin2016domain,li2018deep,arjovsky2019invariant,blanchard2021domain}, and \cite{krueger2021out} as recent exemplars of principled algorithms that are guided by theory and compare their performance with our algorithm's. 

Our proposed upper bound is based on the Wasserstein barycenters. Related to this context are the works \cite{REDKO2017THEORETICAL,SHEN2018WASSERSTEIN}, and \cite{zhou2020domain}. 
In \cite{zhou2020domain}, the pairwise Wasserstein-1 distance \cite{peyre2019computational,OTAM}, is used as a measure of domain discrepancy. Using the dual form of the Wasserstein-1 distance, the feature extractor in \cite{zhou2020domain} minimizes a combination of cross-entropy loss, Wasserstein distance loss, and a contrastive loss to achieve DG. 
The works \cite{REDKO2017THEORETICAL, SHEN2018WASSERSTEIN} provide upper bounds for the risk of unseen domain based on the Wasserstein-1 distance. Although they were originally proposed for DA, they can be adapted to the DG set-up.
While the bounds from \cite{REDKO2017THEORETICAL, SHEN2018WASSERSTEIN} share some similarities with ours, their bounds are constructed in the input space and therefore do not explicitly motivate the use of representation functions. By contrast, our proposed upper bound measures the discrepancy of domains in the representation space, which naturally justifies the decomposition of the hypothesis in the practical implementation. A detailed analysis and comparison of the bounds in \cite{REDKO2017THEORETICAL, SHEN2018WASSERSTEIN} and our proposed bound can be found in Appendix~\ref{apd: compare bounds}.

In addition to the domain-invariant feature learning approach, which is the main focus of this paper, there are other noteworthy and emerging directions in domain generalization research. These are data manipulation techniques \cite{borlino2021rethinking}, meta-learning strategies \cite{du2020learning}, use of pre-trained models \cite{li2022domain}, and seeking flat minima \cite{cha2021swad}. For more details, we refer the reader to \cite{wang2022generalizing,zhou2021domain} which are recent survey articles on DG.

\section{THEORETICAL ANALYSIS AND PROPOSED METHOD}
We consider a \textit{domain} $v$ as a triple $(\mu^{(v)}, f^{(v)}, g^{(v)})$ consisting of a distribution $\mu^{(v)}$ on the input $\bm{x} \in \mathbb{R}^d$, a representation function $f^{(v)}: \mathbb{R}^d \rightarrow \mathbb{R}^{d'}$, from the input space to the representation space, and a stochastic labeling function $g^{(v)}: \mathbb{R}^{d'} \rightarrow \mathcal{Y}$ from the representation space to the label space. We denote the unseen domain by $(\mu^{(u)}, f^{(u)}, g^{(u)})$ and $S$ seen domains by $(\mu^{(s)}, f^{(s)}, g^{(s)})$, with $s = 1, \ldots, S$. 

Let $\mathcal{F} = \{f | f: \mathbb{R}^d \rightarrow \mathbb{R}^{d'} \}$ be the set of \textit{representation functions},  $\mathcal{G}=\{g | g: \mathbb{R}^{d'} \rightarrow \mathcal{Y}\}$ the set of stochastic \textit{labeling functions}, $\mathcal{H} := \mathcal{G} \circ \mathcal{F}$ the set of \textit{hypotheses}, with each hypothesis $h: \mathbb{R}^d \rightarrow \mathcal{Y}$ obtained by composing a $g \in \mathcal{G}$ with an $f \in \mathcal{F}$, \textit{i.e.},  $h = g \circ f$, 
and $\mathcal{D} =\{\psi| \psi: \mathbb{R}^{d'} \rightarrow \mathbb{R}^d \}$ the set of \textit{reconstruction functions} that map from the representation space back to the input space. In this paper, we limit our theoretical study to binary classification problems, specifically hypothesis functions $h$ such that  $h: \mathbb{R}^d \rightarrow \mathcal{Y} =[0,1]$. Note that a similar set-up is also used in \cite{ben2007analysis} where the hypothesis $h$ occurs non-deterministically and maps a data point to a label between zero and one.

The risk of using a hypothesis $h$ in domain $v$ is then defined by:
\begin{equation}
    R^{(v)}(h) := \mathbb{E}_{\bm{x} \sim \mu^{(v)}} \big[\ell(h(\bm{x}),h^{(v)}(\bm{x})) \big],
\end{equation}
where $\mathbb{E}[ \cdot ]$ denotes the expectation, $h^{(v)}=g^{(v)} \circ f^{(v)},$ and $\ell(\cdot,\cdot)$ is a loss function. We make the following assumptions: 
\begin{itemize}
    \item[\textbf{A1:}] The loss function $\ell(\cdot,\cdot)$ is non-negative, symmetric,  bounded by a finite positive number $L$, satisfies the triangle inequality, and $Q$-Lipschitz continuous, \textit{i.e.}, for any three scalars $a,b,c$ and positive constant $Q$, 
    \begin{align}
    |\ell(a,b) -  \ell(a,c)|  \leq Q ~ |b-c|. 
    \end{align}
    
    \item[\textbf{A2:}] The optimal hypothesis of the unseen domain $h^{(u)}=g^{(u)} \circ f^{(u)}$ is $K$-Lipschitz continuous. Specifically, we assume that for any two vectors $\bm{x},\bm{x}' \in \mathbb{R}^d$, and positive constant $K$, 
\begin{align}
| h^{(u)}(\bm{x}) -  h^{(u)}(\bm{x}')| \leq K ~ \| \bm{x} - \bm{x}' \|_2,
\end{align}where $\| \bm{x} - \bm{x}' \|_2$ denotes the Euclidean distance between $\bm{x}$ and $\bm{x}'$.
\end{itemize}

The first four conditions in Assumption A1 can be easily satisfied by any metric or norm truncated by a finite positive number. Concretely, if $d(a,b)$ is a metric, potentially unbounded like Mean Squared Error (MSE), then $loss(a,b) := \min(L, d(a,b))$, where $L$ is a positive constant, will satisfy the first four conditions in A1. The Lipschitz condition in A1 and A2 are also widely used in the theory and practice of DG \cite{blanchard2021domain,SHEN2018WASSERSTEIN,wu2019domain}. 

One may find our assumptions bear some similarities with the assumptions in \cite{REDKO2017THEORETICAL} and \cite{SHEN2018WASSERSTEIN}, but there are some fundamental differences. Specifically, we assume that the loss function is non-negative, symmetric, bounded, Lipschitz, and satisfies the triangle inequality, whereas the loss function in \cite{REDKO2017THEORETICAL} is required to be convex, symmetric, bounded, obey the triangle inequality, and satisfy a specific form. We only assume that the optimal hypothesis function on the unseen domain is Lipschitz, whereas \cite{SHEN2018WASSERSTEIN} requires all hypotheses to be Lipschitz.

\subsection{Bound for Unseen Domain Risk}  \label{sec:bounds}

Our analysis starts by considering a single seen domain. Lemma \ref{LEMMA: 1} below upper bounds the risk $R^{(u)}(h)$ of a hypothesis $h = g \circ f$ in the unseen domain $u$ by four terms: (1) the risk of the seen domain $s$, (2) the $L^1$ distance between the distributions of the \textit{data representations} from the seen and unseen domain, (3) the reconstruction loss that quantifies how well the representation can reconstruct its original data input, and (4) an intrinsic risk term that is free of $h$ and is intrinsic to the domains and the loss function. We use the notation $f_{\#}\mu^{(v)}$ to denote the pushforward of distribution $\mu^{(v)}$ under the representation function $f$, \textit{i.e.}, the distribution of  $f(\bm{x})$ with $\bm{x} \sim \mu^{(v)}$.

\begin{lemma}
\label{LEMMA: 1}
Under assumptions A1 and A2, for any hypothesis $h \in \mathcal{H}$ and any {reconstruction function} $\psi \in \mathcal{D}$, the following bound holds:
\begin{small}
\begin{align*}
    R^{(u)}(h) &\leq R^{(s)}(h) +  L~ \|f_{\#} \mu^{(u)} - f_{\#} \mu^{(s)}\|_1 + QK \Big( \mathbb{E}_{\bm{x} \sim \mu^{(s)}} \big[  \| \psi(f(\bm{x})) - \bm{x} \|_2 \big] 
    + \mathbb{E}_{\bm{x} \sim \mu^{(u)}} \big[ \| \psi (f(\bm{x})) - \bm{x} \|_2 \big] \Big)
    + \sigma^{(u,s)}
\end{align*}\end{small}where 
$\| f_{\#} \mu^{(u)} - f_{\#} \mu^{(s)} \|_1 =\int_{\bm{z}} |f_{\#}\mu^{(u)} - f_{\#}\mu^{(s)}| d \bm{z}$ denotes the $L^1$ distance between $(f_{\#} \mu^{(u)}, f_{\#} \mu^{(s)})$ in the representation space and:
\begin{equation*}
\begin{split}
\sigma^{(u,s)} :=  \min\left\{\ \right. \mathbb{E}_{\bm{x} \sim \mu^{(u)}} \big[\ell(h^{(u)}(\bm{x}),h^{(s)}(\bm{x})) \big],\mathbb{E}_{\bm{x} \sim \mu^{(s)}} \big[\ell(h^{(u)}(\bm{x}),h^{(s)}(\bm{x})) \big]\left. \right\}.
\end{split}
\end{equation*}
\end{lemma}

\begin{proof}
Please see Appendix \ref{apd: proof of Lemma 1 new}.
\end{proof}

In typical DG applications, training data from multiple seen domains are available and can be combined in various ways. Therefore, Lemma \ref{lemma: 2} below extends Lemma \ref{LEMMA: 1} to a convex combination of distributions of multiple seen domains.

\begin{lemma}
\label{lemma: 2} 
For any convex weights $\lambda^{(1)}, \lambda^{(2)}, \ldots, \lambda^{(S)}$ (non-negative and summing to one), any reconstruction function $\psi \in \mathcal{D}$, and any hypothesis $h \in \mathcal{H}$, the following bound holds:
\begin{small}
\begin{align*}
     R^{(u)}(h)  &\leq  \sum_{s=1}^{S} \lambda^{(s)} R^{(s)}(h)  \nonumber\\
     &+  L  \sum_{s=1}^S \lambda ^{(s)}  \|f_{\#} \mu^{(u)} - f_{\#} \mu^{(s)}\|_1 
     \\
     &+ QK \Big( \sum_{s=1}^{S} \lambda^{(s)} \mathbb{E}_{\bm{x} \sim \mu^{(s)}} \big[  \| \psi(f(\bm{x})) - \bm{x} \|_2 \big] + \mathbb{E}_{\bm{x} \sim \mu^{(u)}} \big[ \|\psi (f(\bm{x})) - \bm{x} \|_2 \big] \Big)  \nonumber\\
     &+ \sum_{s=1}^S \lambda^{(s)} \sigma^{(u,s)}.
\end{align*}
\end{small}
\end{lemma}

\begin{proof} {Please see Appendix} \ref{apd: proof of lemma 2-2}.
\end{proof}

The upper bound above relies on the $L^1$ distances between the pushforwards of seen and unseen distributions. However, accurately estimating $L^1$ distances from samples is hard \cite{ben2010theory, kifer2004detecting}. To overcome this practical limitation, we upper bound the $L^1$ distance by the Wasserstein-2 distance under additional regularity assumptions on the pushforward distributions.    

\begin{definition}
\label{def: 1}
\cite{polyanskiy2016wasserstein} A probability distribution on $\mathbb{R}^d$ is called $(c_1,c_2)$-regular, with $c_1, c_2 \geq 0$, if it is absolutely continuous with respect to the Lebesgue measure with a differentiable density $p(\bm{x})$ such that
\[
\forall \bm{x} \in \mathbb{R}^d, \quad \| \nabla \log_2 p(\bm{x}) \|_2 \leq c_1 \| \bm{x} \|_2 + c_2,
\]
where $\nabla$ denotes the gradient and $\| \cdot \|_2$ denotes the Euclidean norm.
\end{definition}

\begin{lemma}
\label{LEMMA: 3}
If $\mu$ and $\nu$ are $(c_1,c_2)$-regular, then:
\begin{small}
\begin{align*}
\|\mu - \nu \|_1  \leq   \sqrt{c_1\Big( \sqrt{\mathbb{E}_{\bm{u} \sim \mu} \big[\|\bm{u} \|_2^2 \big]} + \sqrt{\mathbb{E}_{\bm{v}\sim \nu} \big[\|\bm{v}\|_2^2 \big]} \Big) + 2c_2 } \times \sqrt{\mathsf{W}_2(\mu,\nu)}
\end{align*}\end{small}where the Wasserstein-$p$ metric \cite{peyre2019computational,OTAM} $\mathsf{W}_{p}(\mu, \nu)$ is defined as,
\begin{align*}
 \mathsf{W}_{p}(\mu, \nu) := (\inf_{\pi \in \Pi(\mu, \nu)} \mathbb{E}_{(\bm{u},\bm{v})\sim \pi}[\|\bm{u} - \bm{v}\|_2^p])^{1/p}
\end{align*}
where $\Pi(\mu, \nu)$ is the set of joint distributions with marginals $\mu$ and $\nu$.
\end{lemma}

\begin{proof}
Please see Appendix \ref{apd: proof of Lemma 3}.
\end{proof}

One may wonder what conditions would guarantee the regularity of the pushforward distributions.  Proposition 2 and Proposition 3 in \cite{polyanskiy2016wasserstein} show that any distribution $\nu$ for which $\mathbb{E}_{\bm{v} \sim \nu} \|\bm{v} \|_2$ is finite becomes regular when convolved with any regular distribution, including the Gaussian distribution. Since convolution of distributions corresponds to the addition of independent random vectors having those distributions, it is always possible to make the pushforwards regular by adding a small amount of independent spherical Gaussian noise in the representation space.

Combining Lemma~\ref{lemma: 2}, Lemma~\ref{LEMMA: 3}, and applying Jensen's inequality, we obtain our main result:

\begin{theorem}
\label{THEOREM: 1}
If $f_{\#} \mu^{(s)}$, $s = 1, 2,\dots,S$, and $f_{\#} \mu^{(u)}$ are all $(c_1,c_2)$-regular, then for {any} convex weights $\lambda^{(1)}, \lambda^{(2)}, \ldots, \lambda^{(S)}$, any reconstruction function $\psi \in \mathcal{D}$, and any hypothesis $h \in \mathcal{H}$, the following bound holds:
\begin{small}
\begin{align}
\label{eq: theorem1}
     R^{(u)}(h) &\leq  \sum_{s=1}^S \lambda^{(s)} R^{(s)}(h) \nonumber\\
     &+ LC \big[ \sum_{s=1}^S \lambda^{(s)}  \mathsf{W}^2_2(f_{\#} \mu^{(u)},f_{\#} \mu^{(s)}) \big]^{1/4}  \nonumber \\
     &+ QK \Big( \sum_{s=1}^{S} \lambda^{(s)} \mathbb{E}_{\bm{x} \sim \mu^{(s)}} \big[  \|\psi (f(\bm{x})) - \bm{x} \|_2 \big] + \mathbb{E}_{\bm{x} \sim \mu^{(u)}} \big[ \| \psi (f(\bm{x})) - \bm{x} \|_2 \big] \Big) \nonumber\\
     &+ \sum_{s=1}^S \lambda^{(s)} \sigma^{(u,s)}
\end{align}\end{small}where:
\begin{small}
\begin{equation*}
    C \!=\! \max_{s}  \! \sqrt{c_1 \! \Big(\!  \sqrt{ \mathbb{E}_{\bm{x}\sim \mu^{(u)}}  \! \big[\| f(\bm{x}) \|^2  \big]} \!+\! \sqrt{ \mathbb{E}_{\bm{x}\sim \mu^{(s)}}  \! \big[\| f(\bm{x}) \|^2  \! \big]} \! \Big) \!+\! 2c_2 }.
\end{equation*}
\end{small}
\end{theorem}

\begin{proof}
Please see Appendix \ref{apd: proof of Theorem 1}. 
\end{proof}

The upper bound in Theorem \ref{THEOREM: 1} consists of four terms: the first term is the sum of the risk on seen domains, the second term is the Wasserstein distance between the pushforward of seen and unseen domains in the representation space, the third term indicates how well the input can be reconstructed from its corresponding  representation, and the fourth term is a combined risk that is independent of both the representation function and the labeling function and {only intrinsic} to the domain and loss function. 

The form of the upper bound derived above shares some similarities with previous bounds in \cite{ben2007analysis,REDKO2017THEORETICAL,SHEN2018WASSERSTEIN}. However, it differs from previous bounds in the following important aspects:
\begin{itemize}
    \item Firstly, even though Lemma 1 in \cite{REDKO2017THEORETICAL} and Theorem 1 in \cite{SHEN2018WASSERSTEIN} employ Wasserstein distance to capture domain divergence, the corresponding term is constructed in the \textit{data space}. By contrast, the corresponding term in our bound is constructed in the \textit{representation space}, which not only provides a theoretical justification when decomposing the hypothesis into a representation mapping and a labeling function, but is also consistent with the algorithm implementation in practice. Moreover, the bounds in \cite{REDKO2017THEORETICAL} and \cite{SHEN2018WASSERSTEIN} are controlled by the Wasserstein-$1$ distance, while our upper bound is managed by the square root of the Wasserstein-2 distance. There are regimes where one bound is tighter than the other as discussed in Appendix \ref{apd: compare bounds}. 
\item Secondly, our third term measures how well the input can be reconstructed from its representation. This motivates the use of an encoder-decoder structure in the proposed algorithm in Section \ref{sec:algo} to minimize the reconstruction loss. This is a novel component absent from  \cite{ben2007analysis,REDKO2017THEORETICAL,SHEN2018WASSERSTEIN}.
\item Finally, the last term in our upper bound is independent of both the representation function $f$ and the labeling function $g$. This contrasts with the previous results in \cite{ben2007analysis}, where the last term in their upper bound (see Theorem 1 in \cite{ben2007analysis}) depends on the representation function $f$. We refer the reader to Appendix~\ref{apd: discussion subtlety} for a detailed comparison.
\end{itemize}

The bound proposed in Theorem~\ref{THEOREM: 1} can also be used for the DA problem where one can access the unseen/target domain data and estimate its distribution. However, under the DG setting, the second and third term in  (\ref{eq: theorem1}) are uncontrollable, leading to an intractable upper bound due to the unavailability of the unseen data. This intractability, which cannot be overcome without making additional specific assumptions on the unseen domain, is widely accepted in the literature as a fundamental limitation for all DG methods and analyses.

As a step toward developing a practical algorithm based on our new bound, we decompose both the second term and the third term in  (\ref{eq: theorem1}) into two separate terms where one term completely depends on the unseen distribution and the other fully depends on the seen distributions. 

\begin{corollary}
\label{COR: 1}
Under the setting and notation of Theorem~\ref{THEOREM: 1}, for an arbitrary pushforward distribution $f_{\#} \mu$, we have:
\begin{small}
\begin{align}
 \label{eq: cor1}
     R^{(u)}(h) &\leq  \sum_{s=1}^S \lambda^{(s)} R^{(s)}(h) \nonumber\\
     &+ LC \Big( \sum_{s=1}^S \lambda^{(s)}  \mathsf{W}^2_2(f_{\#} \mu,f_{\#} \mu^{(s)}) \Big)^{1/4} \nonumber\\
     &+ LC \Big( \mathsf{W}^2_2(f_{\#} \mu^{(u)},f_{\#} \mu) \Big)^{1/4} \nonumber\\
     &+ QK \Big( \sum_{s=1}^{S} \lambda^{(s)} \mathbb{E}_{\bm{x} \sim \mu^{(s)}} \big[  \| \psi (f(\bm{x})) - \bm{x} \|_2 \big] \Big) \nonumber\\
     &+ QK \Big( \mathbb{E}_{\bm{x} \sim \mu^{(u)}} \big[ \|\psi (f(\bm{x})) - \bm{x} \|_2 \big] \Big) \nonumber\\
     &+ \sum_{s=1}^S \lambda^{(s)} \sigma^{(u,s)}.
\end{align}
\end{small}
\end{corollary}
\begin{proof}
Please see Appendix \ref{apd: proof of Corollary 1}.
\end{proof}
Motivated by the bound in Corollary \ref{COR: 1}, we want to find a suitable representation function $f$ together with a reconstruction function $\psi$ to minimize the second term $\sum_{s=1}^S \lambda^{(s)} \mathsf{W}^2_2(f_{\#} \mu,f_{\#} \mu^{(s)})$ and the fourth term $\sum_{s=1}^{S} \lambda^{(s)} \mathbb{E}_{\bm{x} \sim \mu^{(s)}} \big[  \|\psi (f(\bm{x})) - \bm{x} \|_2 \big]$ in  (\ref{eq: cor1}), while ignoring the third term $\mathsf{W}^2_2(f_{\#} \mu^{(u)},f_{\#} \mu)$ and the fifth term $\mathbb{E}_{\bm{x} \sim \mu^{(u)}} \big[ \|\psi (f(\bm{x})) - \bm{x} \|_2 \big]$, as both of them are intractable.

Minimizing the second term $\sum_{s=1}^S \lambda^{(s)}  \mathsf{W}^2_2(f_{\#} \mu,f_{\#} \mu^{(s)})$ in  (\ref{eq: cor1}) leads to finding the Wasserstein-2 barycenter of the distributions of seen domains in the representation space.  Here, we assume a uniform weight of $\lambda^{(s)}=\tfrac{1}{S}$ for all $s$, since there is no additional information for selecting these weights.
For this choice, the Wasserstein-2 barycenter of the pushforward distributions of seen domains is defined by:
\begin{equation}
\label{eq: barycenter}
    f_{\#} \mu_{barycenter} := \argmin_{f_{\#} \mu} \sum_{s = 1}^{S} \frac{1}{S} \mathsf{W}_{2}^{2}( f_{\#}\mu^{(s)}, f_{\#} \mu).
\end{equation}
We refer the reader to \cite{agueh2011barycenters,cuturi2014fast} for the definition and properties (existence, uniqueness) of the Wasserstein barycenter.

On the other hand, minimizing the fourth term $\sum_{s=1}^{S} \lambda^{(s)} \mathbb{E}_{\bm{x} \sim \mu^{(s)}} \big[  \|\psi (f(\bm{x})) - \bm{x} \|_2 \big]$ in  (\ref{eq: cor1}) {naturally leads to an auto-encoder mechanism. With a little abuse of notation, we denote the encoder, namely the representation function as $f$ and the decoder, namely the reconstruction function as $\psi$. The $L^2$ reconstruction loss should be optimized over all seen domains.}

\subsection{Proposed Method} \label{sec:methods}
As the last term in (\ref{eq: cor1}) of Corollary \ref{COR: 1} is independent of both the representation function $f$ and the labeling function $g$, and the third and fifth terms are intractable due to their dependence on unseen domain, we focus on designing $f$, $\psi$ and $g$ to minimize the first, second, and fourth terms in (\ref{eq: cor1}) of Corollary \ref{COR: 1}.

Following previous works \cite{albuquerque2019generalizing,ben2010theory, ben2007analysis}, we optimize the first term by training $f$ together with $g$ using a standard cross-entropy (CE) loss, such that the empirical classification risk on seen domains is minimized. The classification loss function can be written as:
\begin{equation}
\label{eq: clf_loss}
\mathsf{L}_{c}(f,g) = \frac{1}{S} \sum_{s=1}^{S} \mathbb{E}_{\bm{x} \sim \mu^{(s)}}[\mathsf{CE}(h^{(s)}(\bm{x}), g (f(\bm{x} )))]
\end{equation}
where $\mathsf{CE}(h^{(s)}(\bm{x}), g (f(\bm{x} )))$
denotes the cross-entropy (CE) loss between the output of classifier and the ground-truth label of seen domain $s$.

As discussed in Corollary \ref{COR: 1}, we propose to use the Wasserstein-2 barycenter of  representation distributions of seen domains to optimize the second term in  (\ref{eq: cor1}). Specifically, the barycenter loss is defined by:
\begin{equation}
\label{eq: baryloss}
    \mathsf{L}_{bary}(f) :=  \sum_{s = 1}^{S} \frac{1}{S} \mathsf{W}_{2}^{2}(f_{\#}\mu^{(s)}, f_{\#} \mu_{barycenter})
\end{equation}
where $f_{\#} \mu_{barycenter}$, as defined in  (\ref{eq: barycenter}), denotes the Wasserstein barycenter of pushforward distributions of seen domains. 

In contrast to the previous Wasserstein distance-based method \cite{zhou2020domain} where pairwise Wasserstein distance loss is employed, we motivate the use of Wasserstein barycenter loss based on our Corollary \ref{COR: 1} and demonstrate its ability in enforcing domain-invariance in the ablation study of Section~\ref{sec:ablation}. Notably, the barycenter loss (\ref{eq: baryloss}) only requires computing $S$ Wasserstein distances, whereas using pairwise Wasserstein distance would require $S(S-1)/2$ computations.

Furthermore, to handle the fourth term in  (\ref{eq: cor1}), we utilize the auto-encoder structure. Specifically, a decoder $\psi: \mathbb{R}^{d'} \rightarrow \mathbb{R}^{d}$ is adopted, leading to the following reconstruction loss term:
\begin{equation}
\label{eq:recon}
        \mathsf{L}_{r}(f,\psi) :=  \frac{1}{S} \sum_{s=1}^{S} \mathbb{E}_{\bm{x} \sim \mu^{(s)}} [\| \bm{x} - \psi(f(\bm{x}))\|^2].
\end{equation}
From the analysis above, {we aim to find a representation function $f$, a classifier $g$, and a decoder function $\psi$ to optimize the following objective function:}
\begin{align}
\label{eq:masterc}
        \arg \min_{f, g, \psi} \mathsf{L}_{c}(f,g) + \alpha \mathsf{L}_{bary}(f) +  \beta  \mathsf{L}_{r}(f,\psi)
\end{align}
where weights $\alpha, \beta > 0$ are hyper-parameters. One can observe that the terms in our proposed upper bound are incorporated into our objective function in (\ref{eq:masterc}). Specifically, the first term in our objective function aims to determine a good classifier $g$ together with a representation mapping $f$ by minimizing the risk of seen domains, which corresponds to the first term of the upper bound in  (\ref{eq: cor1}). The second term in (\ref{eq:masterc}) acts as a domain alignment tool to minimize the discrepancy between seen domains, aligning with the second term in the proposed bound in (\ref{eq: cor1}). 
Note that although $\mathsf{L}_{bary}$ itself requires solving an optimization problem, we leverage fast computation methods, which are also discussed in Section \ref{sec:algo}, to directly estimate this loss without invoking the Kantorovich-Rubenstein dual characterization of Wasserstein distance \cite{OTAM}. This avoids solving a min-max type problem that is often plagued by unstable numerical dynamics. Finally, the third term in the objective function minimizes the mean squared error between the input and its reconstruction over all seen domains, {which directly minimizes the fourth term in  (\ref{eq: cor1})}.

\section{ALGORITHM} \label{sec:algo}

\begin{figure*}[ht]
\begin{center}
\includegraphics[width=0.7\textwidth]{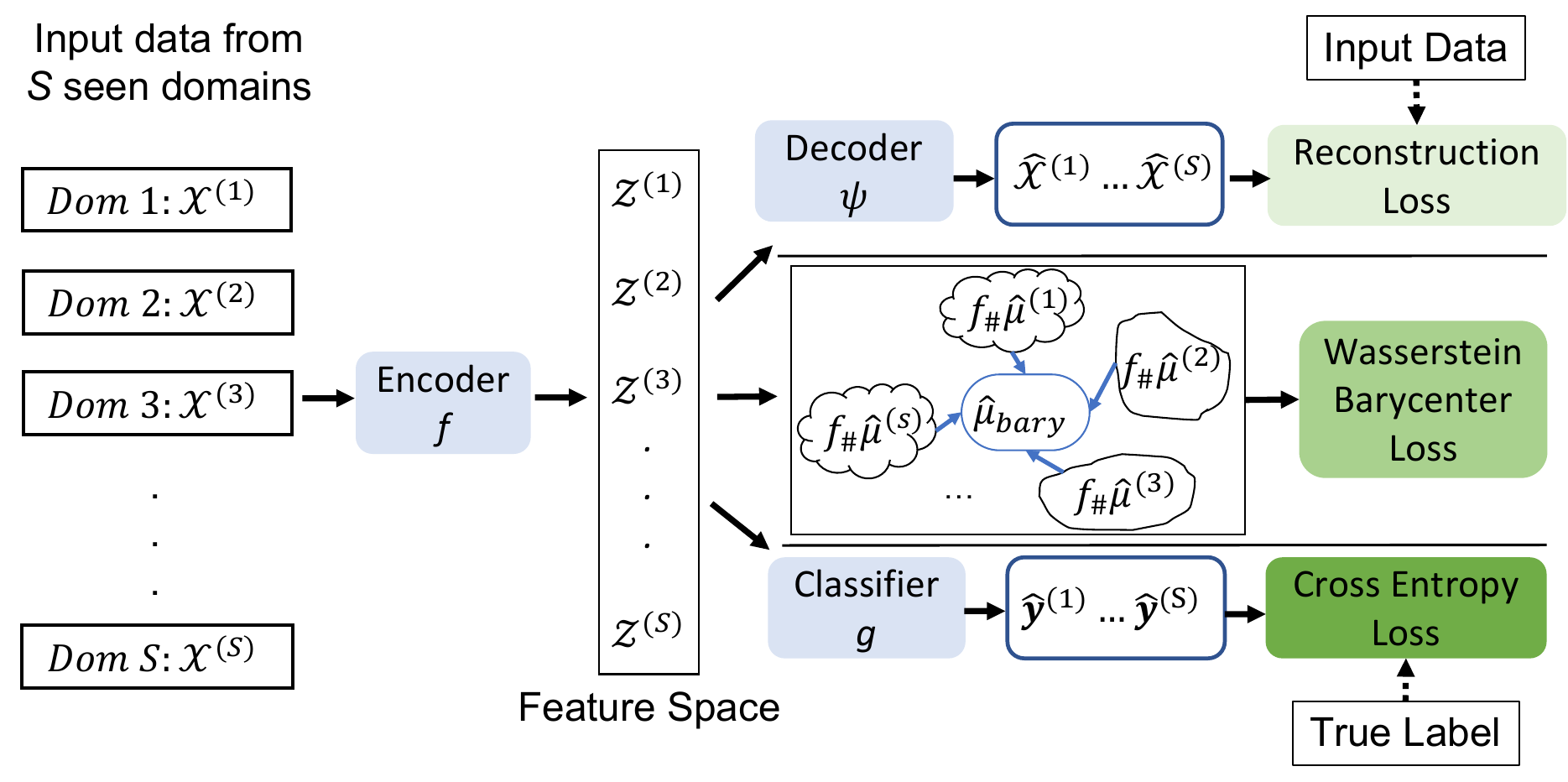}
 \caption{ An overview of the proposed algorithm. The top, middle, and bottom branches refer to the reconstruction loss term, the Wasserstein barycenter loss term, and the classification risk (from seen domains), respectively. }
\label{fig:structure}
\end{center}
\end{figure*}

 Based on the loss function designed above, we propose an algorithm named Wasserstein Barycenter Auto-Encoder (WBAE). The pseudo code of the WBAE algorithm can be found in Algorithm \ref{alg:1} while its block diagram is shown in Fig. \ref{fig:structure}. 

\begin{algorithm}[ht]
    \caption{Wasserstein Barycenter Auto-Encoder (WBAE)}
	\label{alg:1}
	\textbf{Input}: Data from $S$ seen domains, $m$ samples from each domain, learning rate $\eta$, parameters $\alpha,\beta,\epsilon$.
    \textbf{Output}: Encoder $f_{\theta_e}$, decoder $\psi_{\theta_d}$, classifier $g_{\theta_c}$
	\begin{algorithmic}[1]
		\WHILE{training is not end}
		\STATE Randomly choose $m$ samples from each domain, denoted as $\mathcal{X}^{(s)} := \{\bm x_i^{(s)}\}_{i=1}^{m} \sim \hat \mu^{(s)}$ and $\bm y^{(s)} := \{y_i^{(s)}\}_{i=1}^{m}$
		\FOR{$s = 1:S$ and $ i = 1:m$}  
		    \STATE $\bm{z_i}^{(s)}\leftarrow f_{\theta_e}(\bm x_i^{(s)})$ with set $\mathcal{Z}^{(s)} \sim f_{\#}\hat\mu^{(s)}$ 
	    \ENDFOR
		\STATE Calculate the Wasserstein barycenter $\hat \mu_{bary}$ of $\{f_{\#}\hat \mu^{(s)}\}_{s=1}^{S}$ and its supporting points with $f_{\theta_e}$ detached from automatic backpropagation
		\STATE $\mathsf{L}_{bary}\leftarrow \frac{1}{S} \sum_{s=1}^{S} Sinkhorn_{\epsilon}(\hat\mu_{bary}, f_{\#}\hat\mu^{(s)})$
        \STATE $\mathsf{L}_c \leftarrow -\frac{1}{mS}\sum_{s=1}^{S} \sum_{i=1}^{m}y_i^s\log{p(g_{\theta_c}(f_{\theta_e}(\bm x_i^{(s)})))}$
        \STATE $\mathsf{L}_{r} \leftarrow \frac{1}{mS}\sum_{s=1}^{S}\sum_{i=1}^{m}
			 \|\bm{x}_i^{(s)} - \psi_{\theta_d}(\bm{z}_i^{(s)})\|_2^2$
		\STATE $\mathsf{L} \leftarrow \mathsf{L}_c + \alpha \mathsf{L}_{bary} + \beta \mathsf{L}_{r}$
		\STATE $\theta_c \leftarrow \theta_c - \eta \nabla_{\theta_c} \mathsf{L}_c$
		\STATE $\theta_d \leftarrow \theta_d - \eta \nabla_{\theta_d} \mathsf{L}_r$
		\STATE $\theta_e \leftarrow \theta_e - \eta \nabla_{\theta_e} \mathsf{L}$ 
		\ENDWHILE
	\end{algorithmic}
\end{algorithm} 

As shown in the pseudo code, we use an encoder $f$ and a decoder $\psi$, which are parameterized by $\theta_e$ and $\theta_d$ for feature extraction and reconstruction, respectively. Here we denote $\mathcal{X}^{(s)}$ as a set of samples from domain $s$ with empirical distribution $\hat \mu^{(s)}$ and with $\bm x_i^{(s)}$ as one of its element. The corresponding label set of $\mathcal{X}^{(s)}$ is denoted as $\bm y^{(s)}$, where $\bm y^{(s)}:=\{y_i^{(s)}\}$ with $y_i^{(s)}$ as the label for sample $\bm{x}_i^{(s)}$.
The extracted feature $\bm{z_i}^{(s)} = f_{\theta_e}(\bm x_i^{(s)})$ in set $\mathcal{Z}^{(s)}$ follows the empirical distribution of $f_{\#}\hat\mu^{(s)}$. The decoder takes the extracted features as input and outputs the reconstructions as $\psi_{\theta_d}(\bm{z}_i^{(s)})$ for domain $s$. The classifier $g$, parameterized by $\theta_c$ is then applied to the extracted features for label prediction. 

The proposed algorithm requires calculating Wasserstein-2 barycenter and its supporting points. Here we use an off-the-shelf python package \cite{flamary2021pot} that implements a free-support Wasserstein barycenter algorithm described in \cite{cuturi2014fast}. This algorithm is executed in the primal domain and avoids the use of the dual form of Wasserstein distances, which otherwise would turn the problem into an adversarial (min-max) type setting that we want to avoid due to its instability. The barycenter loss is approximated via an average Sinkhorn divergence \cite{feydy2019interpolating} between the seen domains and the estimated barycenter. Sinkhorn divergence is an unbiased proxy for the Wasserstein distance, which leverages entropic regularization \cite{cuturi2013sinkhorn} for computational efficiency, thereby allowing for integrating automatic differentiation with GPU computation. We incorporate the implementation from \cite{feydy2019interpolating} into our algorithm for a fast gradient computation and denote it as $Sinkhorn_{\epsilon}$ in Algorithm \ref{alg:1}, where $\epsilon$ is the entropic regularization term. 

\section{EXPERIMENTS AND RESULTS}\label{sec:eval}

\begin{table}[ht]
\centering
\caption{{Performance of tested methods on PACS dataset in the DomainBed setting, measured by accuracy} (\%). A, C, P, S are left-out unseen domains.}
\vspace{10pt}
\label{table:pacs_res}
\resizebox{0.65\columnwidth}{!}
{\begin{tabular}{lccccc}
\toprule
\textbf{Algorithm}   & \textbf{A}           & \textbf{C}           & \textbf{P}           & \textbf{S}           & \textbf{Avg}         \\
\midrule
\multicolumn{6}{c}{theory-guided algorithms} \\ \midrule
ERM                  & 84.7 $\pm$ 0.4       & 80.8 $\pm$ 0.6       & 97.2 $\pm$ 0.3       & 79.3 $\pm$ 1.0       & 85.5                 \\
IRM                  & 84.8 $\pm$ 1.3       & 76.4 $\pm$ 1.1       & 96.7 $\pm$ 0.6       & 76.1 $\pm$ 1.0       & 83.5                \\
DANN                 & 86.4 $\pm$ 0.8       & 77.4 $\pm$ 0.8       & \textbf{97.3} $\pm$ 0.4       & 73.5 $\pm$ 2.3       & 83.6                \\
CDANN                & 84.6 $\pm$ 1.8       & 75.5 $\pm$ 0.9       & 96.8 $\pm$ 0.3       & 73.5 $\pm$ 0.6       & 82.6                \\
MTL                  & \textbf{87.5} $\pm$ 0.8       & 77.1 $\pm$ 0.5       & 96.4 $\pm$ 0.8       & 77.3 $\pm$ 1.8       & 84.6                \\
VREx                 & 86.0 $\pm$ 1.6       & 79.1 $\pm$ 0.6       & 96.9 $\pm$ 0.5       & 77.7 $\pm$ 1.7       & 84.9                \\
WBAE (Ours) &  86.9 $\pm$ 0.3        & \textbf{81.3} $\pm$ 0.4        & 97.2 $\pm$ 0.2         &  \textbf{80.5} $\pm$ 0.4         & \textbf{86.5}          \\
\midrule
\multicolumn{6}{c}{best-performing heuristic algorithm} \\ \midrule
SagNet               & 87.4 $\pm$ 1.0       & 80.7 $\pm$ 0.6       & 97.1 $\pm$ 0.1       & 80.0 $\pm$ 0.4       & 86.3                 \\
\bottomrule
\end{tabular}}
\end{table}

\begin{table}[ht]
\centering
\caption{{Performance of tested methods on VLCS dataset in the DomainBed setting, measured by accuracy} (\%). C, L, S, V are left-out unseen domains. }
\vspace{10pt}

\label{table:vlcs_res}
\resizebox{0.65\columnwidth}{!}{
\begin{tabular}{lccccc}
\toprule
\textbf{Algorithm}   & \textbf{C}           & \textbf{L}           & \textbf{S}           & \textbf{V}           & \textbf{Avg}         \\
\midrule
\multicolumn{6}{c}{theory-guided algorithms} \\ \midrule
ERM                  & 97.7 $\pm$ 0.4       & 64.3 $\pm$ 0.9       & 73.4 $\pm$ 0.5       & 74.6 $\pm$ 1.3       & 77.5                 \\
IRM                  & 98.6 $\pm$ 0.1       & 64.9 $\pm$ 0.9       & 73.4 $\pm$ 0.6       & 77.3 $\pm$ 0.9       & 78.5               \\
DANN                 & \textbf{99.0} $\pm$ 0.3       & 65.1 $\pm$ 1.4       & 73.1 $\pm$ 0.3       & 77.2 $\pm$ 0.6       & 78.6                 \\
CDANN                & 97.1 $\pm$ 0.3       & 65.1 $\pm$ 1.2       & 70.7 $\pm$ 0.8       & 77.1 $\pm$ 1.5       & 77.5                 \\
MTL                  & 97.8 $\pm$ 0.4       & 64.3 $\pm$ 0.3       & 71.5 $\pm$ 0.7       & 75.3 $\pm$ 1.7       & 77.2                 \\
VREx                 & 98.4 $\pm$ 0.3       & 64.4 $\pm$ 1.4       & \textbf{74.1} $\pm$ 0.4       & 76.2 $\pm$ 1.3       & 78.3                 \\
WBAE (Ours)  &  98.3 $\pm$ 0.2       & 65.5 $\pm$ 1.0        &  72.8 $\pm$ 0.3       &  \textbf{78.6} $\pm$ 0.4   & \textbf{78.8} \\
\midrule
\multicolumn{6}{c}{best-performing heuristic algorithm} \\ \midrule
CORAL                & 98.3 $\pm$ 0.1       & \textbf{66.1} $\pm$ 1.2       & 73.4 $\pm$ 0.3       & 77.5 $\pm$ 1.2       & \textbf{78.8}                  \\

\bottomrule
\end{tabular}}
\end{table}

\begin{table}[ht]
\centering
\caption{{Performance of tested methods on Office-Home dataset in the DomainBed setting, measured by accuracy} (\%). A, C, P, R are left-out unseen domains.}
\vspace{10pt}
\label{table:home_res}
\resizebox{0.65\columnwidth}{!}{
\begin{tabular}{lccccc}
\toprule
\textbf{Algorithm}   & \textbf{A}           & \textbf{C}           & \textbf{P}           & \textbf{R}           & \textbf{Avg}         \\
\midrule
\multicolumn{6}{c}{theory-guided algorithms} \\ \midrule
ERM                  & 61.3 $\pm$ 0.7       & 52.4 $\pm$ 0.3       & 75.8 $\pm$ 0.1       & 76.6 $\pm$ 0.3       & 66.5                 \\
IRM                  & 58.9 $\pm$ 2.3       & 52.2 $\pm$ 1.6       & 72.1 $\pm$ 2.9       & 74.0 $\pm$ 2.5       & 64.3                 \\
DANN                 & 59.9 $\pm$ 1.3       & 53.0 $\pm$ 0.3       & 73.6 $\pm$ 0.7       & 76.9 $\pm$ 0.5       & 65.9                 \\
CDANN                & 61.5 $\pm$ 1.4       & 50.4 $\pm$ 2.4       & 74.4 $\pm$ 0.9       & 76.6 $\pm$ 0.8       & 65.8                 \\
MTL                  & 61.5 $\pm$ 0.7       & 52.4 $\pm$ 0.6       & 74.9 $\pm$ 0.4       & 76.8 $\pm$ 0.4       & 66.4                 \\
VREx                 & 60.7 $\pm$ 0.9       & 53.0 $\pm$ 0.9       & 75.3 $\pm$ 0.1       & 76.6 $\pm$ 0.5       & 66.4                 \\
WBAE (Ours)  &  63.7 $\pm$ 0.5     & \textbf{56.4} $\pm$ 0.8         &  76.1 $\pm$ 0.3        &  \textbf{78.8} $\pm$ 0.4        &  \textbf{68.8}     \\
\midrule
\multicolumn{6}{c}{best-performing heuristic algorithm} \\ \midrule
CORAL     & \textbf{65.3} $\pm$ 0.4       & 54.4 $\pm$ 0.5       & \textbf{76.5} $\pm$ 0.1       & 78.4 $\pm$ 0.5       & 68.7             \\
\bottomrule
\end{tabular}}
\end{table}

\begin{table}[ht]
\centering
\caption{{Performance of tested methods on TerraIncognita dataset in the DomainBed setting, measured by accuracy (\%). L100, L38, L43, L46 are left-out unseen domains.} }
\label{table:terr_res}
\vspace{10pt}
\resizebox{0.65\columnwidth}{!}{
\begin{tabular}{lccccc}
\toprule
\textbf{Algorithm}   & \textbf{L100}           & \textbf{L38}           & \textbf{L43}           & \textbf{L46}           & \textbf{Avg}         \\
\midrule
\multicolumn{6}{c}{theory-guided algorithms} \\ \midrule
ERM                  & 49.8 $\pm$ 4.4       & 42.1 $\pm$ 1.4       & 56.9 $\pm$ 1.8       & 35.7 $\pm$ 3.9       & 46.1                 \\
IRM                  & 54.6 $\pm$ 1.3       & 39.8 $\pm$ 1.9       & 56.2 $\pm$ 1.8       & 39.6 $\pm$ 0.8       & 47.6               \\
DANN                 & 51.1 $\pm$ 3.5       & 40.6 $\pm$ 0.6       & 57.4 $\pm$ 0.5       & 37.7 $\pm$ 1.8       & 46.7                 \\
CDANN                & 47.0 $\pm$ 1.9       & 41.3 $\pm$ 4.8       & 54.9 $\pm$ 1.7       & 39.8 $\pm$ 2.3       & 45.8                 \\
MTL                  & 49.3 $\pm$ 1.2       & 39.6 $\pm$ 6.3       & 55.6 $\pm$ 1.1       & 37.8 $\pm$ 0.8       & 45.6                 \\
VREx                 & 48.2 $\pm$ 4.3       & 41.7 $\pm$ 1.3       & 56.8 $\pm$ 0.8       & 38.7 $\pm$ 3.1       & 46.4                \\
WBAE (Ours)  &  \textbf{55.3} $\pm$ 0.4       & \textbf{44.3} $\pm$ 0.7        &  56.4 $\pm$ 0.5       &  39.1 $\pm$ 0.6   & \textbf{48.8} \\
\midrule
\multicolumn{6}{c}{best-performing heuristic algorithm} \\ \midrule
SagNet               & 53.0 $\pm$ 2.9       & 43.0 $\pm$ 2.5       & \textbf{57.9} $\pm$ 0.6       & \textbf{40.4} $\pm$ 1.3       &48.6                  \\
\bottomrule
\end{tabular}}
\end{table}

\begin{table}[ht]
\setlength{\tabcolsep}{3pt}
\centering
\caption{Performance of theory-guided methods on four datasets in the DomainBed setting, measured by accuracy (\%). The average accuracy is reported over different tasks per dataset.}
\label{table:avg_res}
\vspace{10pt}

\resizebox{0.65\columnwidth}{!}{
\begin{tabular}{lccccc}
\toprule
\textbf{Algorithm}  & \textbf{PACS} & \textbf{VLCS} & \textbf{Office-Home} & \textbf{TerraIncognita} & \textbf{Avg} \\ \midrule
ERM & 85.5 & 77.5 & 66.5 & 46.1 & 68.9 \\
IRM & 83.5 & 78.5 & 64.3 & 47.6 & 68.5 \\
DANN & 83.6 & 78.6 & 65.9 & 46.7 & 68.7 \\
CDANN & 82.6 & 77.5 & 65.8 & 45.8 & 67.9 \\
MTL & 84.6 & 77.2 & 66.4 & 45.6 & 68.5 \\
VREx & 84.9 & 78.3 & 66.4 & 46.4 & 69.0 \\
WBAE (Ours)& \textbf{86.5} & \textbf{78.8} & \textbf{68.8} & \textbf{48.8} & \textbf{70.7} \\ \bottomrule
\end{tabular}}
\end{table}

\begin{table}[ht]
\setlength{\tabcolsep}{3pt}
\centering
\caption{{Performance of tested methods on four datasets in the SWAD setting, measured by accuracy (\%).} }
\vspace{10pt}

\label{table:swad_res}
\resizebox{0.65\columnwidth}{!}{
\begin{tabular}{lccccc}
\toprule
\textbf{Algorithm}   & \textbf{PACS}           & \textbf{VLCS}           & \textbf{Office-Home}           & \textbf{TerraIncognita}           & \textbf{Avg}         \\
\midrule
ERM + SWAD                 & 88.1 $\pm$ 0.1       & 79.1 $\pm$ 0.1       & 70.6 $\pm$ 0.2       & 50.0 $\pm$ 0.3       &  72.0                \\
CORAL + SWAD               & 88.3 $\pm$ 0.1       & 78.9 $\pm$ 0.1       & 71.3 $\pm$ 0.1       & 51.0 $\pm$ 0.1       & 72.4                  \\
WBAE + SWAD &  \textbf{88.4} $\pm$ 0.1       & \textbf{79.5} $\pm$ 0.1        &  \textbf{71.4} $\pm$ 0.2       &  \textbf{51.8} $\pm$ 0.3   & \textbf{72.8} \\
\bottomrule
\end{tabular}}
\end{table}

The proposed method was evaluated on four benchmark datasets for DG: PACS \cite{li2017deeper}, VLCS \cite{fang2013unbiased}, Office-Home \cite{venkateswara2017deep}, and TerraIncognita \cite{beery2018recognition} {under two different settings: DomainBed setting} \cite{gulrajani2020search} {and Stochastic Weight Averaging Densely (SWAD) setting} \cite{cha2021swad}. {In the DomainBed setting, we implemented our method with the widely used DomainBed package and compared it with various \textit{theory-guided} DG algorithms.} 
{Additionally, incorporating the recent advancement in DG-specific optimization, we conducted separate experiments using the SWAD} \cite{cha2021swad} {weight sampling strategy with the same experimental setting described in} \cite{cha2021swad}. {Furthermore, to investigate the impact of different components of the proposed loss function, we conducted an ablation analysis on the PACS, VLCS, and Office-Home datasets and reported the results in} Section~\ref{sec:ablation}.

\subsection{Datasets}

The details for the four datasets are described below:

\begin{itemize}
    \item \textbf{PACS dataset \cite{li2017deeper}:} PACS contains 9,991 images with 7 classes from 4 domains: Art (A), Cartoons (C), Photos (P) and Sketches (S), where each domain represents one type of images.
    \item \textbf{VLCS dataset \cite{fang2013unbiased}:} VLCS consists of {10,729} images from 4 different domains: VOC2007 (V), LabelMe (L), Caltech (C), PASCAL (S). A total of 5 classes are shared by all domains.
    \item \textbf{Office-Home dataset \cite{venkateswara2017deep}:} Office-Home contains 15,500 images from 4 different domains: Artistic (A), Clipart (C), Product (P), and Real-World (R). Each domain has 65 object categories.
    \item \textbf{TerraIncognita dataset\cite{beery2018recognition}:} {TerraIncognita contains four domains \{L100, L38, L43, L46\} with a total of 24,788 pictures of wild animals belonging to 10 classes.}
\end{itemize} 
{Example images of the above datasets are shown in Fig. {\ref{fig:datasets}}, Appendix {\ref{apd: datasets}}.}

\subsection{Methods for Comparison}
In this paper, we compare the empirical performance of our method against the state-of-the-art DG methods reported in \cite{gulrajani2020search} under the DomainBed setting. Specifically, the competing methods include:

\begin{itemize}
    \item Empirical Risk Minimization (\textbf{ERM}) \cite{vapnik1999overview} which aims to minimize the cumulative training error across all seen domains.

    \item Domain-Adversarial Neural Networks (\textbf{DANN}) \cite{ganin2016domain} which is motivated by the theoretical results from \cite{ben2007analysis}. In particular, to minimize the upper bound of the risk in the unseen domain, \textbf{DANN} adopts an adversarial network to enforce that features from different domains are indistinguishable.

    \item Class-conditional DANN (\textbf{C-DANN}) \cite{li2018deep} is a variant of \textbf{DANN} that aims to match the conditional distributions of feature given the label across domains. 
    \item Invariant Risk Minimization (\textbf{IRM}) \cite{arjovsky2019invariant} aims to learn features such that the optimal classifiers applied to these features are matched across domains. 
    
    \item Risk Extrapolation (\textbf{VREx}) \cite{krueger2021out} is constructed on the assumption from \cite{arjovsky2019invariant} which assumes the existence of an optimal linear classifier across all domains. While \textbf{IRM} specifically seeks the invariant classifier, \textbf{VREx} aims to identify the form of the distribution shift and propose a variance penalty, leading to the robustness for a wider variety of distributional shifts.

    \item Marginal Transfer Learning (\textbf{MTL}) \cite{blanchard2011generalizing,blanchard2021domain} is proposed based on an upper bound for the generalization error under the setting of an Agnostic Generative Model. Specifically, \textbf{MTL} estimates the mean embedding per domain and uses it as a second argument for optimizing the classifier.

    \item {CORrelation ALignment} (\textbf{CORAL}) \cite{sun2016deep} is based on the idea of matching the mean and covariance of feature distributions from different domains. 
    
    \item Style-Agnostic Networks (\textbf{SagNet}) \cite{nam2021reducing} minimizes the style induced domain gap by randomizing the style feature for different domains and train the model mainly on the disentangled content feature.
\end{itemize}

We can {categorize the algorithms provided in} \cite{gulrajani2020search} into two groups: (1) heuristic algorithms, {which lack theoretical analysis}, and (2) theory-guided algorithms. {As the proposed method in this paper falls into the second category, we primarily compare it with the theory-guided methods.}
Here, \textbf{ERM} acts as the baseline theory-guided model and \textbf{DANN}, \textbf{C-DANN}, \textbf{IRM}, \textbf{VREx}, \textbf{MTL} are five state-of-the-art theory-guided algorithms. Besides these six methods, for a complete comparison, we also include three heuristic algorithms that achieve the best performances on four evaluated datasets \cite{gulrajani2020search}. More specific, \textbf{SagNet} \cite{nam2021reducing} is the best-performing algorithm for the PACS and {TerraIncognita datasets}, and \textbf{CORAL} \cite{sun2016deep} is the best-performing algorithm for both the VLCS and Office-Home datasets. In the SWAD setting, following \cite{cha2021swad} where \textbf{CORAL} was considered as the representative of the previous state-of-the-art methods, we compare our method with both \textbf{ERM} and \textbf{CORAL}, all of which employed the SWAD strategy. The results for the competing methods above are sourced from \cite{gulrajani2020search} and \cite{cha2021swad}.

\subsection{Experiment Settings}
\textbf{Model Structure}: We used the same feature extractor and classifier as used in \cite{gulrajani2020search} for all four datasets. Specifically, an ImageNet pre-trained ResNet-50 model with the final (softmax) layer removed is used as the feature extractor. The decoder is a stack of 6 ConvTranspose2d layers for all datasets. The detailed structure of the decoder is described in Table \ref{table:decoder_nonmnist}, Appendix ~\ref{appendix:arch_and_hyper}. The classifier is a one-linear-layer model with the output dimension the same as the number of classes.

\noindent\textbf{Hyper-parameters}: {In the DomainBed setting, we performed a random search of 20 trials within the joint distribution of $10^{\text{Uniform}[-3.5, -2]}$ for $\alpha$ and $10^{\text{Uniform}[-3.5, -1.5]}$ for $\beta$} (see (\ref{eq:masterc})) {with other hyper-parameters (\textit{e.g.}, learning rate, batch size, dropout rate, \textit{etc.}) set as the default values recommended in} \cite{gulrajani2020search}. {In the SWAD setting, following} \cite{cha2021swad}, we performed a grid search for $\alpha$ in $\{10^{-3.5}, 10^{-3}, 10^{-2.5}, 10^{-2}\}$ and $\beta$ in $\{10^{-3.5}, 10^{-3}, 10^{-2}, 10^{-1.5}\}$. We chose the value of $\epsilon$ for the Sinkhorn loss (line 7, Algorithm {\ref{alg:1})} as 20, which is the smallest value that can produce stable training processes. A complete description of hyper-parameter tuning in the SWAD setting and a full list of hyper-parameters can be found in Table \ref{table:hyperparameter}, Appendix ~\ref{appendix:arch_and_hyper}.

\noindent\textbf{Model Selection}: We adopted the commonly used training-domain validation strategy in  \cite{gulrajani2020search,krueger2021out} for hyper-parameter tuning and model selection. Specifically, we split the data from each domain into training and validation sets in the proportion 80$\%$ and 20$\%$, respectively. During training, we aggregated together the training/validation samples from each seen domain to form the overall training/validation set and selected the model with the highest validation accuracy for testing.

{All models were trained on a single NVIDIA Tesla V100 16GB GPU.} Experiments on each dataset are repeated three times with different random seeds and the average accuracy together with its standard error are reported. 

\subsection{Results and Ablation Study}
\label{sec:ablation}
{As shown in Table} \ref{table:pacs_res}, \ref{table:vlcs_res}, \ref{table:home_res}, and \ref{table:terr_res}, {the proposed method (WBAE) performs comparably or better than the state-of-the-art methods. In particular, WBAE achieves the highest accuracy in three out of the four datasets compared to all methods, with a moderate improvement over all theory-guided methods on all four datasets. Additionally, the proposed method performs equally well as, or slightly better than, the best-performing heuristic DG methods.} 

In Table \ref{table:pacs_res}, it is demonstrated that WBAE outperforms other theory-guided methods by $0.5\%$ and $1.2\%$ points in both Cartoons (C) and Sketches (S) domains, respectively, and by at least $1\%$ point on average on the PACS dataset. Similarly, Table \ref{table:vlcs_res} {shows that WBAE achieves a performance gain of at least $0.2\%$ points over all theory-guided comparison methods on the VLCS dataset. The effectiveness of the proposed method is further highlighted in Tables} \ref{table:home_res} and \ref{table:terr_res}, {which present results on the larger and more challenging Office-Home and TerraIncognita datasets. Specifically, compared to all theory-guided methods on Office-Home, WBAE boosts the average accuracy by at least $2.3\%$ points on average and at least $2.2\%$, $3.4\%$, $0.3\%$, and $1.9\%$ points on each task. Regarding the TerraIncognita dataset, the proposed algorithm still exhibits superiority by outperforming all theory-guided methods by at least $1.2\%$ points, as shown in Table} \ref{table:terr_res}. {A summary of evaluation results in the DomainBed setting is reported in Table} \ref{table:avg_res}. {The proposed method outperforms all theory-guided methods with a noticeable improvement of $1.7$-$2.8$ percentage points on average across all tested datasets.}

Table \ref{table:swad_res} presents the results obtained by applying SWAD, a DG-specific optimizer and weight-averaging technique, in combination with our proposed algorithm WBAE. It can be observed that this combination outperforms all comparison methods on all four evaluated datasets, with an average improvement of $0.4\%$ point over the previous best-performing method \textbf{CORAL} as reported in \cite{cha2021swad}.

{Based on the results above, it is evident that the proposed algorithm has a more significant impact on the PACS, Office-Home, and TerraIncognita datasets compared to the VLCS dataset.} One possible explanation for this, as also suggested in \cite{zhao2020domain}, is that three out of four domains in the VLCS dataset contain a greater proportion of scenery contents rather than object information. {Unlike the scenery background in TerraIncognita dataset, the scenery contents in the VLCS dataset are usually more intricate and sometimes include multiple objects}, making it challenging for the feature extractor to obtain useful object information for the downstream classification.

\begin{table}[ht]
\centering
\caption{Ablation study for the proposed algorithm (WBAE) on PACS, VLCS, and Office-Home datasets.}
\vspace{10pt}
\label{table:ablation_study}
\resizebox{0.5\columnwidth}{!}{
\begin{tabular}{lccc}
\toprule
Dataset    & no $\mathsf{L}_{bary}$  & no $\mathsf{L}_{r}$   & WBAE            \\ \midrule
PACS        &85.3 $\pm$ 0.3      &86.0 $\pm$ 0.1           &86.5 $\pm$ 0.2 \\
VLCS     &77.9 $\pm$ 0.1  & 78.4 $\pm$ 0.2                   &78.8 $\pm$ 0.2       \\
Office-Home &65.7 $\pm$ 0.2     &67.7 $\pm$ 0.1           & 68.8 $\pm$ 0.1\\ \bottomrule
\end{tabular}}
\end{table}

To study the impact of different components of the loss function in  (\ref{eq:masterc}), we conducted an ablation study for WBAE on {all datasets except TerraIncognita due to our limited computational resources.} In particular, we consider the following variants of our method: (1) no $\mathsf{L}_{bary}$: using the WBAE loss function without the Wasserstein barycenter term $\mathsf{L}_{bary}$; (2) no $\mathsf{L}_r$: using the WBAE loss function without the reconstruction term $\mathsf{L}_r$. We re-ran all the experiments three times using the same model architectures, hyper-parameter tuning, and validation method.

Table~\ref{table:ablation_study} demonstrates the performance of the model with different loss terms removed from the original WBAE loss function. It can be observed that removing $\mathsf{L}_{r}$ from the WBAE loss function leads to a decrease in the accuracy of $0.5\%$, $0.4\%$, and $1.1\%$ points for PACS, VLCS, and Office-Home datasets, respectively. The performance deterioration is more significant when removing $\mathsf{L}_{bary}$ from the WBAE loss function, leading to a drop of $1.2\%$, $0.9\%$, and $3.1\%$ points for PACS, VLCS, and Office-Home datasets, respectively. 
Our ablation study demonstrates the importance of the Wasserstein barycenter loss and also highlights the auxiliary role of the reconstruction loss. Specifically, removing the Wasserstein barycenter loss ($\mathsf{L}_{bary}$) will result in diminished performance, and a similar, though less significant, decrease will occur if the reconstruction loss ($\mathsf{L}_{r}$) is removed.

\section{CONCLUSION AND FUTURE WORK}
\label{sec:conclude}
In this paper, we revisited the theory and methods for DG and provided a new upper bound for the risk in the unseen domain. 
The proposed upper bound contains four terms: (1) the empirical risk of the seen domains in the input space; (2) the discrepancy between the induced representation distribution of seen and unseen domains, which can be further represented by the Wasserstein-2 barycenter of representation in the seen domains; (3) the reconstruction loss term that measures how well the data can be reconstructed from its representation; and (4) a combined risk term. The proposed upper bound provides valuable insights in three aspects. Firstly, we observed that the combined risk term in previous bounds relied on the representation function, which made optimization challenging. By contrast, our combined risk term in the proposed upper bound is a constant with respect to both the representation and the labeling function, making optimization straightforward, thus bridging the previous gap between theory and practice.
Secondly, compared with other upper bounds using Wasserstein distance to measure the domain discrepancy, the proposed bound constructs the discrepancy term in the representation space rather than in the data space. This approach offers a theoretical justification for the decomposition of the hypothesis when bounding the risk and for practical implementation when designing the algorithm. 
Lastly, motivated by the proposed upper bound, our practical algorithm WBAE demonstrates competitive performance over state-of-the-art DG algorithms, validating the usefulness of the proposed theoretical bound for addressing the DG problem. In addition, our bound encourages minimizing the reconstruction loss term, which theoretically supports the use of (nearly) invertible representation mappings in recent works \cite{johansson2019support,nguyen2022trade}.
In terms of algorithm and numerical implementation, it should be noted that while our theory-guided method is effective in addressing the DG problem, it may become computationally expensive if one wants to use a larger batch size for a more accurate estimation of the Wasserstein-2 barycenter. 
To alleviate this constraint, our future works will focus on leveraging the recently proposed large-scale-barycenter and mapping estimators \cite{fan2020scalable,korotin2022wasserstein} {to enable the calculation of barycenters with a larger number of samples.}

\section*{Acknowledgments} 
This work was supported in part by the Air Force Office of Scientific Research under award number FA9550-18-1-0465 and NSF CAREER award CCF:1553075, NSF RAISE 1931978, NSF ERC planning 1937057.

\newpage
\appendix
\section{Limitations of Previous Upper Bounds}
\label{apd: discussion subtlety}
First, let us recall that a \textit{domain} $v$ is defined as a triple $(\mu^{(v)}, f^{(v)}, g^{(v)})$ consisting of a distribution $\mu^{(v)}$ on the input $\bm{x} \in \mathbb{R}^d$, a representation function $f^{(v)}: \mathbb{R}^d \rightarrow \mathbb{R}^{d'}$ that maps an input $\bm{x}$ from the input space to its representation $\bm{z}$ in the representation space, and a stochastic labeling function $g^{(v)}: \mathbb{R}^{d'} \rightarrow \mathcal{Y}$ that maps the representation space $\mathbb{R}^{d'}$ to a label space $\mathcal{Y}$.

We denote the unseen domain by $(\mu^{(u)}, f^{(u)}, g^{(u)})$ and the seen domain by $(\mu^{(s)}, f^{(s)}, g^{(s)})$. Let $\mathcal{F} = \{f | f: \mathbb{R}^d \rightarrow \mathbb{R}^{d'} \}$ be a set of \textit{representation functions},  $\mathcal{G}=\{g | g: \mathbb{R}^{d'} \rightarrow \mathcal{Y}\}$ a set of stochastic \textit{labeling functions}. Here, the label space $\mathcal{Y}$ is considered as binary. A \textit{hypothesis} $h: \mathbb{R}^d \rightarrow \mathcal{Y}$ is obtained by composing each $g \in \mathcal{G}$ with each $f \in \mathcal{F}$, \textit{i.e.},  $h = g \circ f$.  Next, we rewrite Theorem 1 in \cite{ben2007analysis} using our notations.

\noindent\textbf{Theorem 1 in  \cite{ben2007analysis}}. \textit{Let $f$ be a fixed representation function from the input space to representation space and $\mathcal{G}$ be a hypothesis space of
VC-dimension $k$. If random labeled samples of size $m$ are generated by applying $f$ to i.i.d. samples from the seen domain, then with probability at least $1- \delta$, for every $g \in \mathcal{G}$:
\begin{align}
     R^{(u)}(g) & \leq  R^{(s)}(g)+d_{\mathcal{H}}(f_{\#} \mu^{(u)}, f_{\#} \mu^{(s)})+ \lambda \label{bound in [4]}\\
     &\leq \hat{R}^{(s)}(g) + \sqrt{\dfrac{4}{m} \big( k \log \dfrac{2em}{k}+\log\dfrac{4}{\delta} \big)} 
     + d_{\mathcal{H}}(f_{\#} \mu^{(u)}, f_{\#} \mu^{(s)}) + \lambda
\end{align}
where $e$ is the base of the natural logarithm, $d_{\mathcal{H}}$ is $\mathcal{H}$-divergence (please see Definition 1 in \cite{ben2010theory}, Definition 2.1 in \cite{zhao2019learning} or Definition 1 in \cite{kifer2004detecting}), $R^{(u)}(g)=\mathbb{E}_{\bm{z} \sim f_{\# }\mu^{(u)}}\big[| g(\bm{z}) -g^{(u)}(\bm{z})|\big]$ denotes the risk in the unseen domain, $R^{(s)}(g)=\mathbb{E}_{\bm{z} \sim f_{\#} \mu^{(s)}}\big[| g(\bm{z}) -g^{(s)}(\bm{z})|\big]$ and $\hat{R}^{(s)}(g)$ denote the risk in the seen domain and its empirical estimation, respectively, and: 
\begin{equation}
\label{eq: combined risk}
\lambda = \inf_{g \in \mathcal{G}} \big( R^{(s)}(g)+R^{(u)}(g) \big)
\end{equation}
is the \textit{combined risk}.}

Although the bound in Theorem 1 of \cite{ben2007analysis} was originally constructed for the domain adaptation problem, it has significantly influenced past and recent works in domain generalization as discussed earlier in Section~\ref{sec:intro}. To highlight the differences between our work and previous theoretical bounds (the bound in Theorem 1 of \cite{ben2007analysis} and Theorem 4.1 of \cite{zhao2019learning}), we provide a detailed comparison below:  

\begin{itemize}
    \item Firstly, \cite{ben2007analysis} defines the risk induced by labeling function $g$ from the representation space to the label space based on the disagreement between $g$ and the optimal labeling function $g^{(u)}$:
    \begin{equation}
    \label{eq: risk defined in [4]}
      R^{(u)}(g)=\mathbb{E}_{\bm{z} \sim f_{\#} \mu^{(u)}}\big[| g(\bm{z}) -g^{(u)}(\bm{z})|\big].  
    \end{equation}

On the other hand, we define the risk induced by using a hypothesis $h$ from the input space to the label space by the disagreement between $h$ and the optimal hypothesis $h^{(u)}$ via a general loss function $\ell(\cdot,\cdot)$:
\begin{equation}
\label{eq: risk defined in our paper}
    R^{(u)}(h) = \mathbb{E}_{\bm{x} \sim \mu^{(u)}} \big[\ell(h(\bm{x}),h^{(u)}(\bm{x})) \big].
\end{equation}
Since the empirical risk measures the probability of misclassification of a hypothesis that maps from the input space to the label space, minimizing $R^{(u)}(g)$ does not guarantee to minimize the empirical risk. Though there are some cases for the causality to hold, for example, if the representation function $f$ is invertible \textit{i.e.}, there is a one-to-one mapping between $\bm{x}$ and $\bm{z}$, and the loss function has the form of $\ell(a,b)=|a-b|$, it is possible to verify that $R^{(u)}(g)=R^{(u)}(h)$. In general, the representation mapping might not be invertible. For example, let us consider a representation function $f$ that maps $f(\bm{x}_1)=f(\bm{x}_2)=\bm{z}$, $\bm{x_1} \neq \bm{x_2}$, with corresponding labels as $y_1=0$ and $y_2=1$. In this case, the risk defined in (\ref{eq: risk defined in [4]}) will introduce a larger error than the risk introduced in (\ref{eq: risk defined in our paper}) since $g(\bm{z})$ cannot be mapped to both ``0'' and ``1''. That said, the risk defined in (\ref{eq: risk defined in our paper}) is more precise to describe the empirical risk. In addition, the risk defined in (\ref{eq: risk defined in [4]}) is only a special case of (\ref{eq: risk defined in our paper}) when the representation mapping $f$ is invertible and the loss function satisfies $\ell(a,b)=|a-b|$.

\item Secondly, using the setting in \cite{ben2007analysis}, for a given hypothesis space, the ideal joint hypothesis $g^*$ is defined as the hypothesis which globally minimizes the combined error from seen and unseen domains \cite{ben2007analysis, ben2010theory}:
$$g^*= \argmin_{g \in \mathcal{G}}\big( R^{(s)}(g)+R^{(u)}(g) \big).$$
In other words, this hypothesis should work well in both domains. The error induced by using this ideal joint hypothesis is called \textit{combined risk}:
 $$\lambda = \inf_{g \in \mathcal{G}} \big( R^{(s)}(g)+R^{(u)}(g) \big) = \big( R^{(s)}(g^*)+R^{(u)}(g^*) \big).$$ 
Note that the labeling function $g$ is a mapping from the representation space to the label space, therefore, the ideal labeling function $g^*$ depends implicitly on the representation function $f$, hence,  $\lambda$ depends on $f$. Simply ignoring this fact and treating $\lambda$ as a constant may loosen the upper bound. By contrast, our goal is to construct an upper bound with the \textit{combined risk} term $\sigma^{(u,s)}$ independent of both the representation function and the labeling function, which can be seen from Lemma \ref{LEMMA: 1} and Theorem \ref{THEOREM: 1}.

\item Finally, it is worth comparing our upper bound with the bound in Theorem 4.1 of \cite{zhao2019learning} which also has the \textit{combined risk} term free of the choice of the hypothesis class. However, note that the result in Theorem 4.1 of \cite{zhao2019learning} does not consider any representation function $f$, \textit{i.e.}, their labeling function directly maps from the input space to the label space, while our hypothesis is composed of a representation function from the input space to the representation space followed by a labeling function from the representation space to the label space. Since it is possible to pick a representation function $f$ that maps any input to itself, \textit{i.e.}, $f(\bm{x})=\bm{x}$ which leads to $h=g \circ f = g$, the bound in \cite{zhao2019learning} can be viewed as a special case of our proposed upper bound in Lemma~\ref{LEMMA: 1}.
\end{itemize}

\section{Comparison with Upper Bounds in \cite{REDKO2017THEORETICAL} and \cite{SHEN2018WASSERSTEIN}}
\label{apd: compare bounds}

The form of the proposed upper bound derived in Theorem~\ref{THEOREM: 1} shares some similarities with Lemma 1 in \cite{REDKO2017THEORETICAL} and Theorem 1 in \cite{SHEN2018WASSERSTEIN}, for example, all of them introduce Wasserstein distance between domain distributions. However, they differ in the following key aspects.

\begin{enumerate}
\setlength \itemsep{0 pt}
    \item The term containing Wasserstein distance in our upper bound is constructed in the \textit{representation} space, not in the data (ambient) space, which provides a theoretical justification when decomposing the hypothesis into a representation mapping and a labeling function. This is also consistent with the algorithmic implementation in practice.

    \item The bounds in Lemma 1 of \cite{REDKO2017THEORETICAL} and Theorem 1 of \cite{SHEN2018WASSERSTEIN} are controlled by the Wasserstein-$1$ distance  while our upper bound is managed by the square-root of the Wasserstein-2 distance. There are regimes where one bound is tighter than the other. It is well-known that $\mathsf{W}_1(\mu,\nu) \leq \mathsf{W}_2(\mu,\nu)$, if $\mathsf{W}_2(\mu,\nu) \leq 1$, then $\mathsf{W}_1(\mu,\nu) \leq \sqrt{\mathsf{W}_2(\mu,\nu)}$. However, based on Jensen's inequality, it is possible to show that $\sqrt{\mathsf{W}_2(\mu,\nu)} \leq [Diam(f(\bm{X})) \mathsf{W}_1(\mu,\nu)]^{1/4}$ where $Diam(f(\bm{X}))$ denotes the largest distance between two points in the representation space $\mathbb{R}^{d'}$ generated by input $\bm{X}$ via mapping $f$. To guarantee $\sqrt{\mathsf{W}_2(\mu,\nu)} \leq \mathsf{W}_1(\mu,\nu)$, a sufficient condition is $[Diam(f(\bm{X})) \mathsf{W}_1(\mu,\nu)]^{1/4} \leq \mathsf{W}_1(\mu,\nu)$ which is equivalent to $Diam(f(\bm{X})) \leq \mathsf{W}_1(\mu,\nu)^3$. In fact, for a given $Diam(f(\bm{X}))$, the larger the value of $\mathsf{W}_1(\mu,\nu)$, the higher the chance that this sufficient condition will hold.
\end{enumerate}

\section{Proof of Lemma \ref{LEMMA: 1}}
\label{apd: proof of Lemma 1 new}

Note that in this paper, we assume that any hypothesis function $h(\cdot)$ outputs a value in $[0,1]$, \textit{i.e.}, $h: \mathbb{R}^d \rightarrow [0,1]$, and $\ell(\cdot, \cdot)$ is a bounded distance metric. In addition, we assume that $h^{(u)}(\cdot)$ is $K$-Lipschitz continuous and $\ell(\cdot)$ is $Q$-Lipschitz continuous. Particularly, we assume that for any two vectors $\bm{x}, \bm{x}' \in \mathbb{R}^d$ and any three scalars  $a,b,$ and $c$, the following inequalities hold:
\begin{align}
| h^{(u)}(\bm{x}) -  h^{(u)}(\bm{x}')|  \leq K \|\bm{x}-\bm{x}'\|_2, 
\end{align}
\begin{align}
|\ell(a,b) -  \ell(a,c)|  \leq Q |b-c|, 
\end{align}where $\|\bm{x} - \bm{x}'\|_2$ and $|b - c|$ denote the Euclidean distances between $\bm{x}$ and $\bm{x}'$, and $b$ and $c$, respectively. 

\noindent \textbf{Lemma 2.1.} 
If $h^{(u)}(\cdot)$ is $K$-Lipschitz continuous and $\ell(\cdot)$ is $Q$-Lipschitz continuous. Then, for any hypothesis $h \in \mathcal{H}$ and any function (decoder) $\psi : \mathbb{R}^{d'} \rightarrow \mathbb{R}^d$, the following bound holds:
\begin{align*}
    R^{(u)}(h) &\leq R^{(s)}(h) +  L~ \|f_{\#} \mu^{(u)} - f_{\#} \mu^{(s)}\|_1 \nonumber\\
    &+ QK \Big( \mathbb{E}_{\bm{x} \sim \mu^{(s)}} \big[  \|\psi (f(\bm{x})) - \bm{x} \|_2 \big] \nonumber\\
    &+ \mathbb{E}_{\bm{x} \sim \mu^{(u)}} \big[ \|\psi (f(\bm{x})) - \bm{x} \|_2 \big] \Big)
    + \sigma^{(u,s)}
\end{align*}where 
$\| f_{\#} \mu^{(u)} - f_{\#} \mu^{(s)} \|_1 =\int_{\bm{z}} |f_{\#}\mu^{(u)} - f_{\#}\mu^{(s)}| d \bm{z}$ denotes the $L^1$ distance between $(f_{\#} \mu^{(u)}, f_{\#} \mu^{(s)})$ and:
\begin{equation*}
\begin{split}
\sigma^{(u,s)} :=  \min\left \{ \right. &\mathbb{E}_{\bm{x} \sim \mu^{(u)}} \big[\ell(h^{(u)}(\bm{x}),h^{(s)}(\bm{x})) \big],\mathbb{E}_{\bm{x} \sim \mu^{(s)}} \big[\ell(h^{(u)}(\bm{x}),h^{(s)}(\bm{x})) \big]\left. \right \}.
\end{split}
\end{equation*}

\begin{proof}
First, we want to note that our approach  is motivated by the proof of Theorem 1 in \cite{ben2010theory}. Next, to better demonstrate the relationship between the hypothesis, input distribution, true representation and labeling functions, we use inner product notation $\langle \cdot , \cdot \rangle$ to {denote} expectations. Specifically,  
\begin{equation}
    R^{(v)}(h) := \mathbb{E}_{\bm{x} \sim \mu^{(v)}} \big[\ell(h(\bm{x}),h^{(v)}(\bm{x})) \big]=\langle \ell(h,h^{(v)}),\mu^{(v)} \rangle.
\end{equation}
From the definition of risk,
\begin{small}
\begin{align}
\label{eq: nguyen 69}
   R^{(u)}(h) 
   &= \langle \ell(h,h^{(u)}),\mu^{(u)} \rangle \nonumber\\
   &=  \langle  \ell(h,h^{(s)}), \mu^{(s)}  \rangle - \langle  \ell(h,h^{(s)}),  \mu^{(s)}  \rangle + \langle  \ell(h,h^{(u)}),\mu^{(u)} \rangle \nonumber\\
   &= R^{(s)}(h) + \big( \langle \ell(h,h^{(u)}),\mu^{(u)} \rangle - \langle \ell(h,h^{(s)}),\mu^{(u)} \rangle \big)
   + \big( \langle \ell(h,h^{(s)}),\mu^{(u)} \rangle - \langle \ell(h,h^{(s)}),\mu^{(s)} \rangle \big) \nonumber\\
   &\leq R^{(s)}(h) + \langle \ell(h^{(u)},h^{(s)}),\mu^{(u)} \rangle
   + \langle \ell(h,h^{(s)}),\mu^{(u)}-\mu^{(s)} \rangle
\end{align}
\end{small}where the inequality of (\ref{eq: nguyen 69}) follows from the triangle inequality $\ell(h,h^{(u)}) \leq \ell(h,h^{(s)})+ \ell(h^{(s)},h^{(u)})$ and $\ell(h^{(s)},h^{(u)})=\ell(h^{(u)},h^{(s)})$. 

In an analogous fashion, it is possible to show that:
\begin{small}
\begin{align}
\label{eq: nguyen 70}
    R^{(u)}(h)   \leq R^{(s)}(h) + \langle \ell(h^{(u)},h^{(s)}),\mu^{(s)} \rangle 
     + \langle \ell(h,h^{(u)}),\mu^{(u)}-\mu^{(s)} \rangle.
\end{align}\end{small}

Next, we will bound the third term in the right-hand-side of (\ref{eq: nguyen 70}). Specifically,

\begin{small}
\begin{align}
&\langle  \ell(h,h^{(u)}),\mu^{(u)}-\mu^{(s)} \rangle \nonumber\\
&  = \mathbb{E}_{\bm{x} \sim \mu^{(u)}} \Big[ \ell \big(h(\bm{x}),h^{(u)}(\bm{x}) \big) \Big] - \mathbb{E}_{\bm{x} \sim \mu^{(s)}} \Big[ \ell \big(h(\bm{x}),h^{(u)}(\bm{x}) \big) \Big] \nonumber \\
&  \leq  \! \max \! \Big\{ \!\mathbb{E}_{\bm{x} \sim \mu^{(u)}} \! \Big[ \ell  \big( h(\bm{x}),h^{(u)}(\psi (f(\bm{x}))) \!+\! K \|\psi (f(\bm{x})) \!-\! \bm{x} \|_2 \! \big) \! \Big],
\mathbb{E}_{\bm{x} \sim \mu^{(u)}} \! \Big[\! \ell \big(  h(\bm{x}),h^{(u)}(\psi (f(\bm{x}))) \!-\! K \|\psi (f(\bm{x})) \!-\! \bm{x} \|_2 \!\big) \! \Big] \! \Big\} \nonumber\\
& - \min \! \Big\{ \! \mathbb{E}_{\bm{x} \sim \mu^{(s)}} \! \Big[ \ell \big( h(\bm{x}),h^{(u)}( \psi (f(\bm{x}))) \!+\! K \| \psi (f(\bm{x})) \!-\! \bm{x} \|_2 \! \big) \! \Big],
\mathbb{E}_{\bm{x} \sim \mu^{(s)}} \! \Big[ \ell \big(\! h(\bm{x}),h^{(u)}(\psi (f(\bm{x}))) \!-\! K \|\psi (f(\bm{x})) \!-\! \bm{x} \|_2 \!\big) \! \Big]  \! \Big \} \label{eq: an2} \\
& \leq  \Big( \mathbb{E}_{\bm{x} \sim \mu^{(u)}} \Big[ \ell \big( h(\bm{x}),h^{(u)}(\psi (f(\bm{x})))  \big) \Big] 
 +  \mathbb{E}_{\bm{x} \sim \mu^{(u)}} \Big[Q K \|\psi (f(\bm{x})) - \bm{x} \|_2  \Big] \Big) \nonumber\\ 
& \hspace{10pt} - \Big( \mathbb{E}_{\bm{x} \sim \mu^{(s)}} \Big[ \ell \big( h(\bm{x}),h^{(u)}(\psi (f(\bm{x}))) \big) \Big] 
- \mathbb{E}_{\bm{x} \sim \mu^{(s)}} \Big[Q K \| \psi (f(\bm{x})) - \bm{x} \|_2  \Big] \Big) \label{eq: an4} \\
&  = \Big( \mathbb{E}_{\bm{z} \sim f_{\#}\mu^{(u)}} \Big[ \ell \big( g(\bm{z}),h^{(u)} (\psi(\bm{z})) \big) \Big] 
 - \mathbb{E}_{\bm{z} \sim f_{\#}\mu^{(s)}} \Big[ \ell \big( g(\bm{z}),h^{(u)} (\psi(\bm{z})) \big) \Big] \Big) \nonumber \\
&\hspace{10pt}  + \Big( \mathbb{E}_{\bm{x} \sim \mu^{(u)}} \Big[ Q K \|\psi(f(\bm{x})) - \bm{x} \|_2 \Big] 
+ \mathbb{E}_{\bm{x} \sim \mu^{(s)}} \Big[ Q K \|\psi(f(\bm{x})) - \bm{x} \|_2 \Big] \Big) \label{eq: an3}\\ 
& = \langle \ell \big( g (\bm{z}),h^{(u)} (\psi (\bm{z})) \big), f_{\#} \mu^{(u)} - f_{\#} \mu^{(s)} \rangle 
+ QK \!\Big(\! \mathbb{E}_{\bm{x} \sim \mu^{(s)}} \big[ \|\psi(f(\bm{x})) \!-\! \bm{x} \|_2 \! \big] \!+\! \mathbb{E}_{\bm{x} \sim \mu^{(u)}} \big[ \|\psi(f(\bm{x})) \!-\! \bm{x} \|_2 \! \big] \! \Big) \nonumber \\
& \leq   L \langle 1,  |f_{\#} \mu^{(u)} - f_{\#} \mu^{(s)}| \rangle 
+  QK  \Big(\! \mathbb{E}_{\bm{x} \sim \mu^{(s)}} \big[ \|\psi(f(\bm{x})) \!-\! \bm{x} \|_2 \!\big] \!+\! \mathbb{E}_{\bm{x} \sim \mu^{(u)}} \! \big[ \|\psi(f(\bm{x})) \!-\! \bm{x} \|_2 \! \big] \! \Big) \label{eq: nguyen 75}.
\end{align}
\end{small}

Note that in this paper, any hypothesis function $h(\cdot)$ is assumed to output a scalar value in $[0,1]$, \textit{i.e.}, $h: \mathbb{R}^d \rightarrow [0,1]$, and $\ell(\cdot, \cdot)$ is a distance metric. With all these assumptions, we get (\ref{eq: an2}) due to $\min \{\ell(a,c), \ell(a,d)\} \leq \ell(a,b) \leq \max\{ \ell(a,c), \ell(a,d)\}$, $\forall b \in [c,d], a, b, c, d \in \mathbb{R}$ and the fact for Lipschitz function $h^{(u)}(\cdot)$ that:
\begin{align}
 \label{eq: lipschitz for hu}
    h^{(u)}(\psi(f(\bm{x}))) - K \|\psi(f(\bm{x})) - \bm{x} \|_2  \leq  h^{(u)}(\bm{x}),  
\end{align}
\begin{align}
 \label{eq: lipschitz for hu1}
 h^{(u)}(\bm{x}) \leq  h^{(u)}(\psi(f(\bm{x}))) +  K \|\psi(f(\bm{x})) - \bm{x} \|_2.   
\end{align}

Next, (\ref{eq: an4}) is due to the Lipschitzness of $\ell(\cdot)$:
\begin{small}
\begin{align}
 \label{eq: lipschitz for ell}
 \max \Big\{ & \ell \Big( h(\bm{x}),h^{(u)}(\psi(f(\bm{x}))) + K \|\psi(f(\bm{x})) - \bm{x} \|_2 \Big), 
\ell \Big( h(\bm{x}),h^{(u)}(\psi(f(\bm{x}))) - K \|\psi(f(\bm{x})) - \bm{x} \|_2 \Big)  \Big\} \nonumber\\
\leq \, &  \ell \Big( h(\bm{x}),h^{(u)}(\psi(f(\bm{x}))) \Big) + Q K \|\psi(f(\bm{x})) - \bm{x} \|_2,
\end{align}
\begin{align}
 \label{eq: lipschitz for ell 2}
\min \Big\{ & \ell \Big( h(\bm{x}),h^{(u)}(\psi(f(\bm{x}))) + K \|\psi(f(\bm{x})) - \bm{x} \|_2 \Big), 
 \ell \Big( h(\bm{x}),h^{(u)}(\psi(f(\bm{x}))) - K \|\psi(f(\bm{x})) - \bm{x} \|_2 \Big)  \Big\} \nonumber\\
\geq \, &  \ell \Big( h(\bm{x}),h^{(u)}(\psi(f(\bm{x}))) \Big) - Q K \|\psi(f(\bm{x})) - \bm{x} \|_2.
\end{align}\end{small}

Finally, we get (\ref{eq: an3}) due to $h=g \circ f$, $f(\bm{x})=\bm{z}$, and (\ref{eq: nguyen 75}) due to $\ell(\cdot, \cdot)$ is bounded by $L$.  

The proof of Lemma follows by combining (\ref{eq: nguyen 69}), (\ref{eq: nguyen 70}), (\ref{eq: nguyen 75}), and  note that: 
$$\sigma^{(u,s)} = \min \Big\{ \langle \ell(h^{(u)},  h^{(s)}), \mu^{(u)}  \rangle, \langle \ell(h^{(u)},  h^{(s)}), \mu^{(s)}  \rangle \Big\},$$ and $$\langle 1,  |f_{\#} \mu^{(u)}-f_{\#} \mu^{(s)}| \rangle= \|f_{\#} \mu^{(u)}-f_{\#} \mu^{(s)}\|_1.$$
\end{proof}

\section{Proof of Lemma II.2 }
\label{apd: proof of lemma 2-2}

Apply Lemma \ref{LEMMA: 1} $S$ times for $S$ seen domains, then for any hypothesis $h \in \mathcal{H}$ and function (decoder) $\psi : \mathbb{R}^{d'} \rightarrow \mathbb{R}^d$, the following bound holds: 
\begin{small}
\begin{align}
\label{eq: just add 1}
    R^{(u)}(h) &\leq R^{(s)}(h) +  L~ \|f_{\#} \mu^{(u)} - f_{\#} \mu^{(s)}\|_1 \nonumber\\
    &+ QK \Big( \mathbb{E}_{\bm{x} \sim \mu^{(s)}} \big[  \|\psi (f(\bm{x})) - \bm{x} \|_2 \big] \nonumber\\
    &+ \mathbb{E}_{\bm{x} \sim \mu^{(u)}} \big[ \|\psi (f(\bm{x})) - \bm{x} \|_2 \big] \Big)
    + \sigma^{(u,s)}, \forall s=1,\dots,S.
\end{align}\end{small}{Next, multiplying} (\ref{eq: just add 1}) {with its corresponding convex weight $\lambda^{(s)}$, for $s=1,2,\dots,S$, and summing them up, we have:}
\begin{small}
\begin{align}
\label{eq: just add 2}
    \sum_{s=1}^S \lambda^{(s)} R^{(u)}(h) &\leq \sum_{s=1}^S \lambda^{(s)} \Bigg[  R^{(s)}(h) +  L~ \|f_{\#} \mu^{(u)} - f_{\#} \mu^{(s)}\|_1 
    + QK \Big( \mathbb{E}_{\bm{x} \sim \mu^{(s)}} \big[  \|\psi (f(\bm{x})) - \bm{x} \|_2 \big] \nonumber\\
    &+ \mathbb{E}_{\bm{x} \sim \mu^{(u)}} \big[ \|\psi (f(\bm{x})) - \bm{x} \|_2 \big] \Big)
    + \sigma^{(u,s)} \Bigg].
\end{align}\end{small}Note that $\sum_{s=1}^S \lambda^{(i)}=1$, thus, the left-hand side of (\ref{eq: just add 2}) {is $R^{(u)}(h)$, and by re-arranging the terms on the right-hand side of} (\ref{eq: just add 2}), the proof follows.

\section{Proof of Lemma \ref{LEMMA: 3}}
\label{apd: proof of Lemma 3}

From Pinsker's inequality \cite{csiszar2011information}, the $L^1$ distance can be bounded by Kullback–Leibler (KL) divergence as follows: 
\begin{equation}
\label{eq: pinsker's inequality}
    \| \mu-\nu \|_1^2 \leq 2 d_{KL}(\mu,\nu)
\end{equation}
where $\| \mu-\nu \|_1$ and $d_{KL}(\mu,\nu)$ denote $L^1$ distance and Kullback–Leibler divergence between two distributions $\mu$ and $\nu$, respectively. Since $\| \mu-\nu \|_1=\| \nu - \mu\|_1$, applying Pinsker's inequality to $(\mu,\nu)$ and $(\nu,\mu)$, 
\begin{equation}
    2 \| \mu-\nu \|_1^2 = \| \mu-\nu \|_1^2 + \| \nu-\mu \|_1^2  \leq 2 d_{KL}(\mu,\nu) + 2 d_{KL}(\nu,\mu)
\end{equation}
which is equivalent to,
\begin{equation}
\label{eq: thuan 3-a}
\| \mu - \nu \|_1 \leq  \sqrt{d_{KL}(\mu,\nu) + d_{KL}(\nu,\mu)}.
\end{equation}

Next, if $\mu$ and $\nu$ are $(c_1,c_2)$-regular distributions, their Kullback–Leibler divergences can be bounded by their Wasserstein-2 distance as follows (please see equation (10), Proposition 1 in \cite{polyanskiy2016wasserstein}),
\begin{small}
\begin{align}
\label{eq: KL to WS}
&d_{KL}(\mu,\nu) + d_{KL}(\nu,\mu) 
\leq 2\mathsf{W}_2(\mu,\nu)\Big(
\dfrac{c_1}{2}  
\sqrt{\mathbb{E}_{\bm{u}\sim \mu}\big[\|\bm{u}\|_2^2\big]} + 
\dfrac{c_1}{2}
\sqrt{\mathbb{E}_{\bm{v}\sim \nu}\big[\|\bm{v}\|_2^2\big]}  +c_2\Big).
\end{align}
\end{small}

Combining (\ref{eq: thuan 3-a}) and (\ref{eq: KL to WS}), we have:
\begin{small}
\begin{align}
\label{eq: thuan 49}
    &\| \mu - \nu \|_1 
    \leq \! \big[\mathsf{W}_2(\mu,\nu) \big]^{1/2} \! \sqrt{c_1\Big( 
    \sqrt{\mathbb{E}_{\bm{u}\sim \mu}\big[\|\bm{u}\|_2^2\big]} \!+\! 
    \sqrt{\mathbb{E}_{\bm{v}\sim \nu}\big[\|\bm{v}\|_2^2\big]}
    \Big) \!+\! 2c_2}. 
\end{align}
\end{small}

\section{Proof of Theorem \ref{THEOREM: 1}}
\label{apd: proof of Theorem 1}

Under the assumption that $f_{\#} \mu^{(s)}$ and $f_{\#} \mu^{(u)}$ are $(c_1,c_2)$-regular, $\forall s=1,2,\dots,S$, we can derive the following inequality from Lemma \ref{LEMMA: 3},
\begin{small}
\begin{align}
 &\|f_{\#} \mu^{(u)} - f_{\#} \mu^{(s)}\|_1 \leq 
  \sqrt{c_1
  \Big( 
  \sqrt{\mathbb{E}_{\bm{x}\sim \mu^{(s)}}\big[\|f(\bm{x})\|_2^2\big]}
  +
  \sqrt{\mathbb{E}_{\bm{x}\sim \mu^{(u)}}\big[\|f(\bm{x})\|_2^2\big]}  
  \Big) 
  + 
  2c_2 } 
  \times \big[\mathsf{W}_2(f_{\#} \mu^{(u)},f_{\#} \mu^{(s)}) \big]^{1/2}. \label{eq: nguyen 100}
\end{align}
\end{small}
Let:
\begin{small}
\begin{equation}
C \!:=\! \max_{s} \!  
\sqrt{\! c_1 \!
  \Big( \!
  \! \sqrt{\mathbb{E}_{\bm{x}\sim \mu^{(s)}} \! \big[ \! \|f(\bm{x})\|_2^2 \! \big]}
  \!+\! 
  \sqrt{ \! \mathbb{E}_{\bm{x}\sim \mu^{(u)}} \! \big[ \! \|f(\bm{x})\|_2^2 \! \big]}  
  \! \Big) 
  \!+\! 
  2c_2 }.
\end{equation}
\end{small}Multiplying (\ref{eq: nguyen 100}) by $\lambda^{(s)}$ and summing over all $s$, we get:
\begin{small}
\begin{equation}
\label{eq: nguyen 102}
    \sum_{s=1}^S \! \lambda^{(s)} \!   \|f_{\#} \mu^{(u)} \!-\! f_{\#} \mu^{(s)}\|_1 \!\leq\! C \! \sum_{s=1}^S \! \lambda^{(s)} \! \big[\mathsf{W}_2(f_{\#} \mu^{(u)}, \!f_{\#} \mu^{(s)}) \! \big]^{1/2}. 
\end{equation}
\end{small}
By Jensen's inequality,
\begin{small}
\begin{equation}
\label{eq: nguyen 104}
   \sum_{s=1}^S \! \lambda^{(s)} \! \big[\! \mathsf{W}_2( \! f_{\#} \! \mu^{(u)} \!, \! f_{\#} \! \mu^{(s)}) \! \big]^{1/2}  \!\leq\!    \big[ \! \sum_{s=1}^S \! \lambda^{(s)}  \mathsf{W}^2_2( \! f_{\#} \! \mu^{(u)} \!, \!f_{\#} \! \mu^{(s)}) \! \big]^{1/4}.
\end{equation}
\end{small}
From (\ref{eq: nguyen 102}) and (\ref{eq: nguyen 104}),
\begin{small}
\begin{equation}
\label{eq: nguyen 106}
\sum_{s=1}^S \! \lambda^{(s)}   \|f_{\#} \! \mu^{(u)} \!-\! f_{\#} \! \mu^{(s)}\|_1   \!\leq \!  C  \! \big[ \! \sum_{s=1}^S \! \lambda^{(s)}  \mathsf{W}^2_2(f_{\#} \! \mu^{(u)},f_{\#} \! \mu^{(s)}) \! \big]^{1/4}.
\end{equation}
\end{small}Finally, combining the upper bound in Lemma \ref{lemma: 2} and (\ref{eq: nguyen 106}), the proof follows.

\section{Proof of Corollary \ref{COR: 1}}
\label{apd: proof of Corollary 1}

We begin with the second term in the upper bound of Theorem \ref{THEOREM: 1}. Indeed, for any arbitrary pushforward distribution $f_{\#} \mu$, we have:
\begin{small}
\begin{align}
 & \Big[ \sum_{s=1}^S \lambda^{(s)}  \mathsf{W}^2_2(f_{\#} \mu^{(u)},f_{\#} \mu^{(s)}) \Big]^{1/4} \\
\leq& \Big[ \sum_{s=1}^S \lambda^{(s)} \Big( \mathsf{W}^2_2(f_{\#} \mu^{(u)},f_{\#} \mu ) + \mathsf{W}^2_2(f_{\#} \mu ,f_{\#} \mu^{(s)}) \Big) \Big]^{1/4} \label{eq: corolary 11}\\
=& \! \Big[\! \sum_{s=1}^S \! \lambda^{(s)}  \mathsf{W}^2_2(f_{\#} \! \mu^{(u)},f_{\#} \! \mu) \!+\! \sum_{s=1}^S \! \lambda^{(s)}  \mathsf{W}^2_2(f_{\#} \! \mu,f_{\#} \! \mu^{(s)}) \! \Big]^{1/4} \\
=& \Big[ \mathsf{W}^2_2(f_{\#} \mu^{(u)},f_{\#} \mu) + \sum_{s=1}^S \lambda^{(s)}  \mathsf{W}^2_2(f_{\#} \mu,f_{\#} \mu^{(s)}) \Big]^{1/4} \label{eq: corolary 12}\\
\leq& \! \Big[ \! \sum_{s=1}^S \lambda^{(s)}  \mathsf{W}^2_2(f_{\#} \mu,f_{\#} \mu^{(s)}) \! \Big]^{1/4} \!+\! \Big[\! \mathsf{W}^2_2(f_{\#} \mu^{(u)},f_{\#} \mu) \! \Big]^{1/4}  
\label{eq: corolary 13}
\end{align}
\end{small}with (\ref{eq: corolary 11}) due to the triangle inequality, (\ref{eq: corolary 12}) due to $ \sum_{s=1}^S \lambda^{(s)}=1$, (\ref{eq: corolary 13}) due to the fact that for any $a,b \geq 0$ and $0<p \leq 1$, $(a+b)^{p} \leq a^p + b^p$.

Combining (\ref{eq: theorem1}) in Theorem \ref{THEOREM: 1} and  (\ref{eq: corolary 13}), the proof of Corollary \ref{COR: 1} follows.

\section{EXAMPLE IMAGES OF FOUR TESTED DATASETS}
\label{apd: datasets}
\begin{figure*}[ht]
\centering
\subfigure[PACS]{\includegraphics[width=0.2\textwidth]{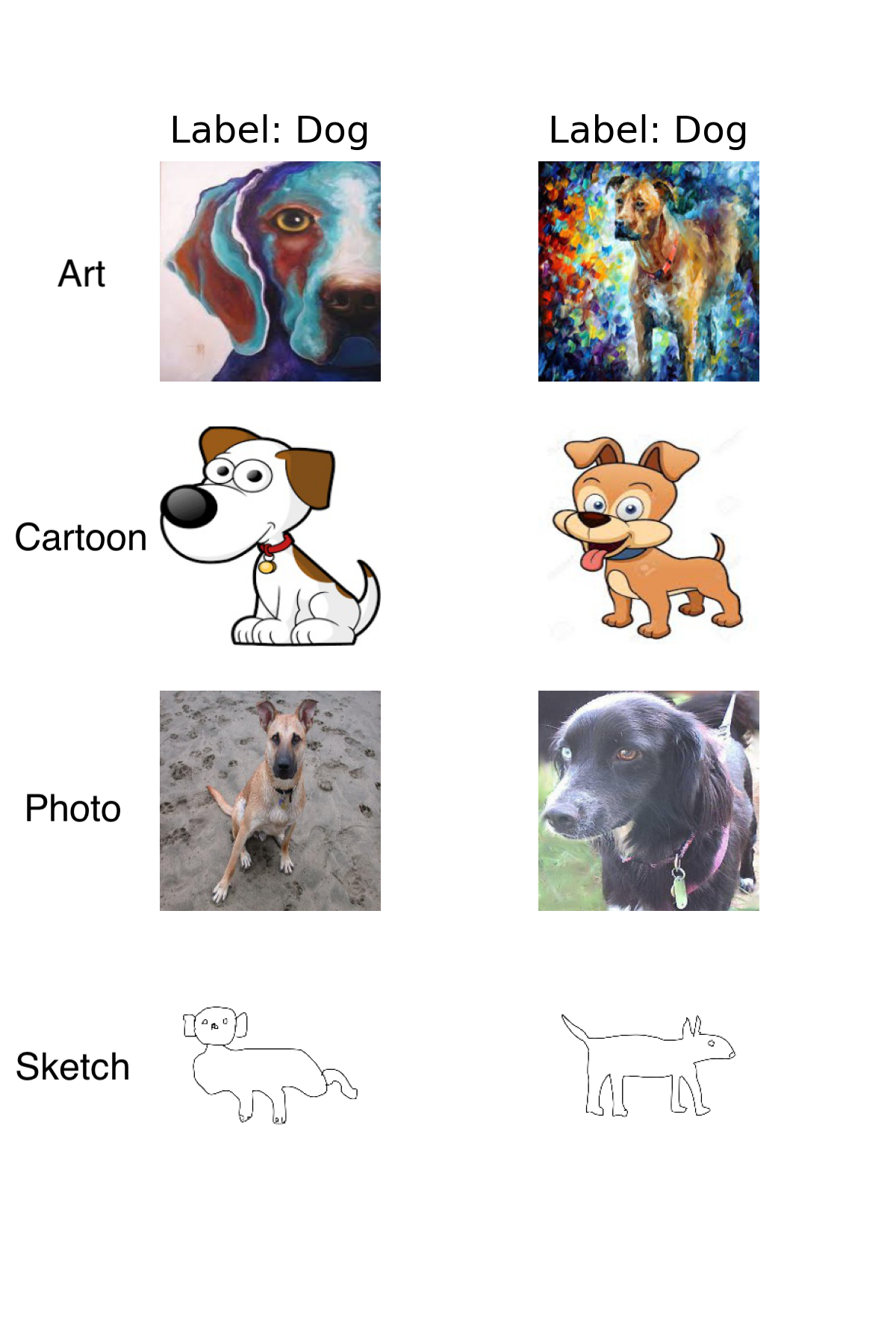}}%
\subfigure[VLCS]{\includegraphics[width=0.2\textwidth]{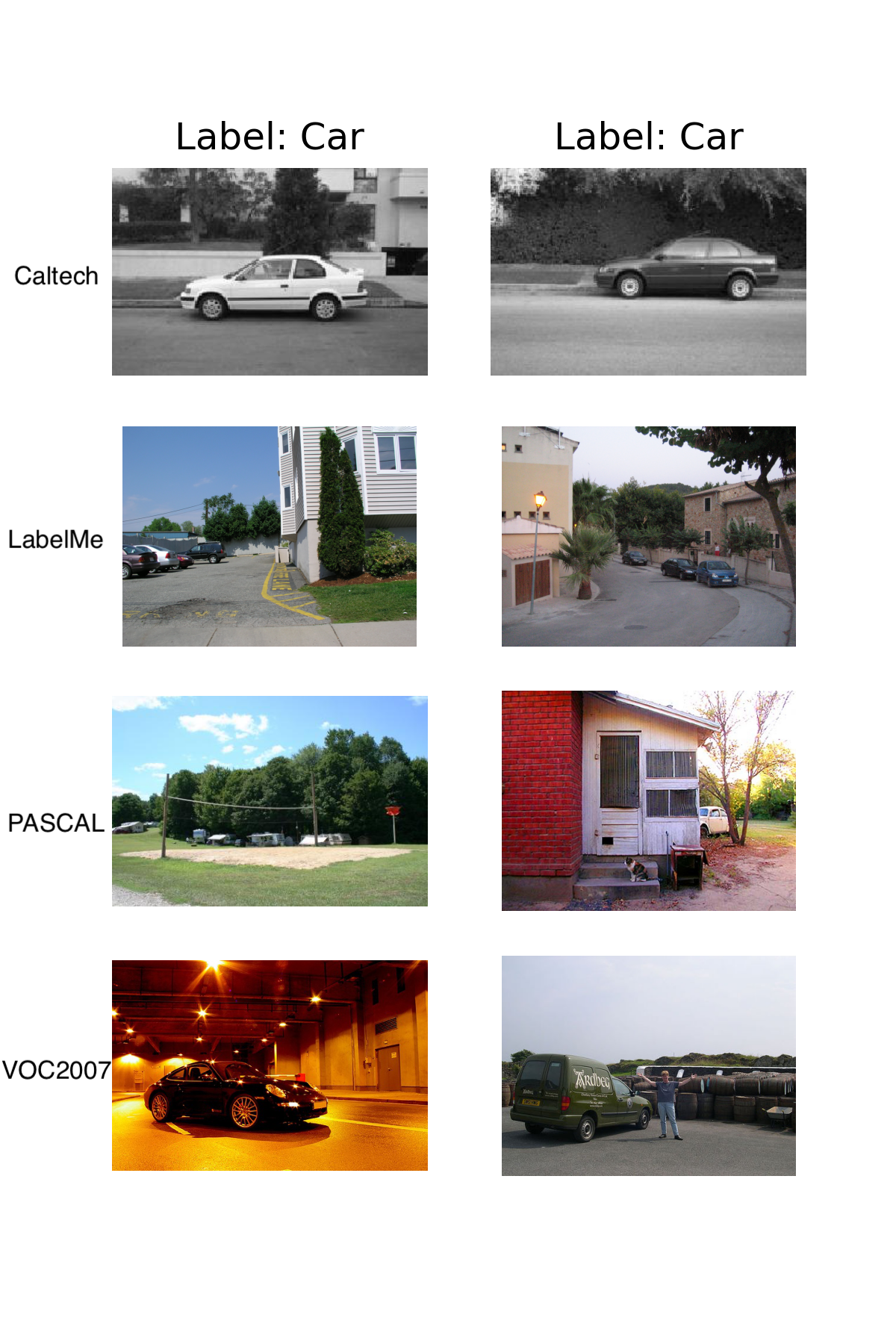}}%
\subfigure[Office-Home]{\includegraphics[width=0.2\textwidth]{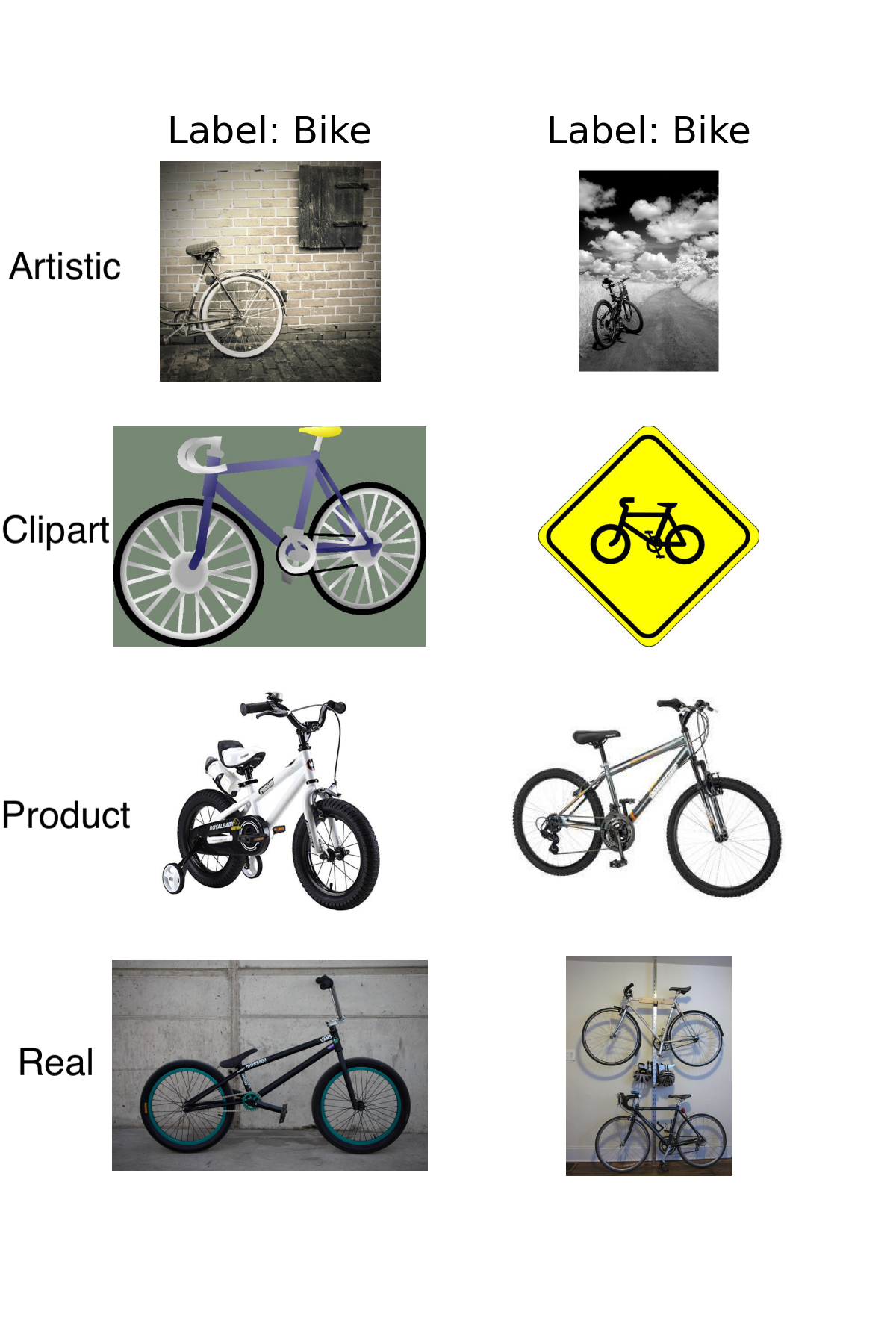}}%
\subfigure[TerraIncognita]{\includegraphics[width=0.2\textwidth]{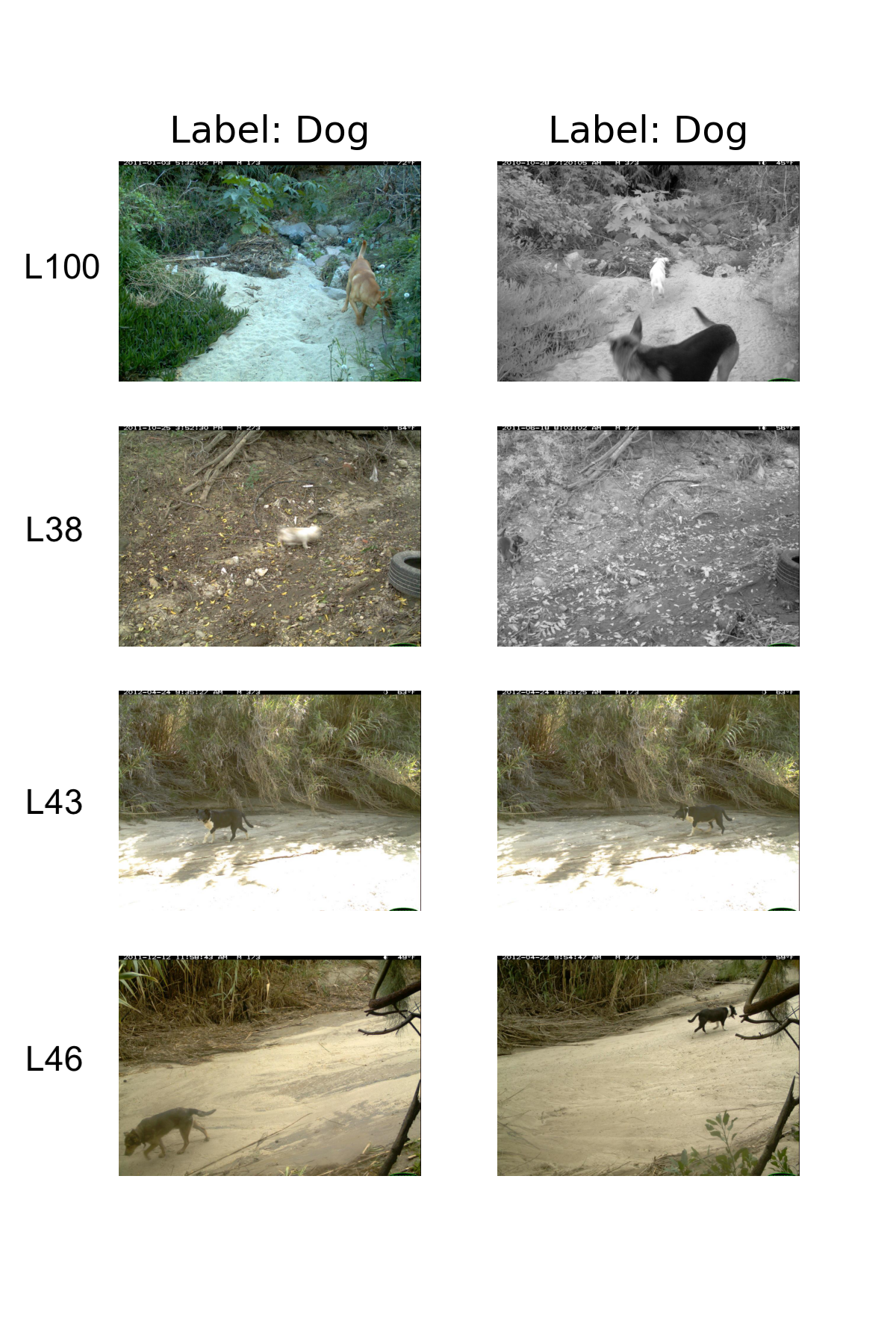}}%
\caption{{Example images of four tested datasets.}}
\label{fig:datasets}
\end{figure*}

Example images of each dataset are shown in Fig. \ref{fig:datasets}.

\section{ARCHITECTURE AND HYPER-PARAMETERS}
\label{appendix:arch_and_hyper}

\begin{itemize}
    \item The model structure of the decoder used in {all four} datasets can be found in Table \ref{table:decoder_nonmnist}. 
    
    \item A list of hyper-parameters used in our proposed method is shown in Table \ref{table:hyperparameter}. 
\end{itemize}

\begin{table}[ht]
\caption{Model structure of the decoder.}
\centering
\vspace{10pt}
\label{table:decoder_nonmnist}
\resizebox{0.65\columnwidth}{!}{
\begin{tabular}{l}
\toprule
Layer                                                                                      \\ \midrule
ConvTranspose2d (in=2048, out=512, kernel\_size=4, stride= 1, padding=0) \\
BatchNorm2d  + ReLU                                                                        \\
ConvTranspose2d (in=512, out=256, kernel\_size=4, stride=2, padding=1)  \\
BatchNorm2d  + ReLU                                                                        \\
ConvTranspose2d (in=256, out=128, kernel\_size=4, stride=2, padding=1)   \\
BatchNorm2d  + ReLU                                                                        \\
ConvTranspose2d (in=128, out=64, kernel\_size=4, stride=2, padding=1)    \\
BatchNorm2d  + ReLU                                                                        \\
ConvTranspose2d (in=64, out=32, kernel\_size=4, stride=2, padding=1)     \\
BatchNorm2d  + ReLU                                                                        \\
ConvTranspose2d (in=32, out=3, kernel\_size=4, stride=2, padding=1)      \\
Tanh + Interpolate (size=(224, 224))                                                       \\ \bottomrule
\end{tabular}}
\end{table}

\begin{table}[ht]
\centering
\caption{Hyper-parameters of the proposed method.}
\vspace{10pt}
\label{table:hyperparameter}
\resizebox{0.65\columnwidth}{!}{
\begin{tabular}{@{}lll@{}}
\toprule
Parameters     & DomainBed Setting                     & SWAD Setting        \\ \midrule
Optimizer & Adam\cite{KingmaB14}                & Adam \cite{KingmaB14}\\   
Learning rate  & $5 \times 10^{-5}$                            & \{$10^{-5}$, $3 \times 10^{-5}$, $5 \times 10^{-5}$\}                          \\
Batch size     & 32                                & 32                                \\
ResNet dropout & 0                                 & \{0.0, 0.1, 0.5\}                   \\
Weight decay   & 0                                 & \{$10^{-4}$, $10^{-6}$\}                                 \\
Training steps & 2000                              & 5000                              \\
$\epsilon$     & 20                                & 20                                \\
$\alpha$       & $10^{\text{Uniform}(-3.5, -2)}$   & $\{10^{-3.5}, 10^{-3}, 10^{-2.5}, 10^{-2}\}$   \\
$\beta$        & $10^{\text{Uniform}(-3.5, -1.5)}$ &  $\{10^{-3.5}, 10^{-3}, 10^{-2}, 10^{-1.5}\}$\\ \bottomrule
\end{tabular}}
\end{table}

Following \cite{cha2021swad}, in the SWAD setting, we first fixed all algorithm-agnostic hyper-parameters (HPs) and only tuned the algorithm-specific HPs. Specifically, we first fixed the learning rate as $5 \times 10^{-5}$, Resnet dropout rate and weight decay both as 0, and grid searched  $\alpha, \beta$ in $\{10^{-3.5}, 10^{-3}, 10^{-2.5}, 10^{-2}\}$ and $\{10^{-3.5}, 10^{-3}, 10^{-2}, 10^{-1.5}\}$ with the batch size as 32. Then we searched learning rate, Resnet dropout rate, and weight decay in  \{$10^{-5}$, $3 \times 10^{-5}$, $5 \times 10^{-5}$\}, \{$0.0$, $0.1$, $0.5$\} and \{$10^{-4}$, $10^{-6}$\}, with the selected $\alpha, \beta$, as performed in \cite{cha2021swad}. 
We used the same values for SWAD-specific hyper-parameters as those used in \cite{cha2021swad}, without any further tuning.
\section{Code Availability}
The code used to generate the results and tables is available in the GitHub repository:
\url{https://github.com/boyanglyu/DG_via_WB}.

\bibliography{refs}
\end{document}